\documentclass[journal,twoside,web]{ieeecolor}
\usepackage{tmi}
\usepackage{cite}
\usepackage{ulem}
\usepackage{color}
\usepackage{float}
\usepackage{amsmath,amssymb,amsfonts}
\usepackage{algorithmic}
\usepackage{graphicx}
\usepackage{textcomp}
\usepackage{subfigure}
\usepackage{algorithm}
\usepackage{multirow}
\usepackage{booktabs}
\def\BibTeX{{\rm B\kern-.05em{\sc i\kern-.025em b}\kern-.08em
    T\kern-.1667em\lower.7ex\hbox{E}\kern-.125emX}}
\begin{document}
\title{DeepACEv2: Automated Chromosome Enumeration in Metaphase Cell Images Using Deep Convolutional Neural Networks}
\author{Li Xiao, Chunlong Luo, Tianqi Yu, Yufan Luo, Manqing Wang, Fuhai Yu, Yinhao Li, Chan Tian, Jie Qiao
\thanks{Copyright (c) 2019 IEEE. Personal use of this material is permitted. However, permission to use this material for any other purposes must be obtained from the IEEE by sending a request to $pubs-permissions@ieee.org$.}
\thanks{ Li Xiao and Chunlong Luo contribute equally to this work. This work was supported by National Natural Science Foundation of China(grant 31900979) to Li Xiao. Editorial corresponding author and Lead author(PI of this project): Li Xiao. (Email: {\color{blue}xiaoli@ict.ac.cn}) Corresponding Author:(Li Xiao, Jie Qiao, Chan Tian)}
\thanks{Li Xiao is with Advanced Computer Research Center and MIRACLE group, Key Laboratory of Intelligent Information Processing Institute of Computing Technology, Chinese Academy of Science and Ningbo HuaMei Hospital, University of the Chinese Academy of Sciences (UCAS). (email: xiaoli@ict.ac.cn)}
\thanks{Chunlong Luo,  Yufan Luo, Yinhao Li are students with Advanced Computer Research Center, Key Laboratory of Intelligent Information Processing Institute of Computing Technology, Chinese Academy of Science and School of Computer and Control Engineering, University of the Chinese Academy of Sciences (UCAS).}
\thanks{Tianqi Yu, Manqing Wang, Fuhai Yu, Chan Tian, and Jie Qiao are with Reproductive Medicine Center, Peking University Third Hospital. (email: jie.qiao@263.net,tianchan\_cdc@126.com).}}
\maketitle

\begin{abstract}
Chromosome enumeration is an essential but tedious procedure in karyotyping analysis. To automate the enumeration process, we develop a chromosome enumeration framework, DeepACEv2, based on the region based object detection scheme. The framework is developed following three steps. Firstly, we take the classical ResNet-101 as the backbone and attach the Feature Pyramid Network (FPN) to the backbone. The FPN takes full advantage of the multiple level features, and we only output the level of feature map that most of the chromosomes are assigned to. Secondly, we enhance the region proposal network's ability by adding a newly proposed Hard Negative Anchors Sampling to extract unapparent but essential information about highly confusing partial chromosomes. Next, to alleviate serious occlusion problems, besides the traditional detection branch, we novelly introduce an isolated Template Module branch to extract unique embeddings of each proposal by utilizing the chromosome's geometric information. The embeddings are further incorporated into the No Maximum Suppression (NMS) procedure to improve the detection of overlapping chromosomes. Finally, we design a Truncated Normalized Repulsion Loss and add it to the loss function to avoid inaccurate localization caused by occlusion. In the newly collected $1375$ metaphase images that came from a clinical laboratory, a series of ablation studies validate the effectiveness of each proposed module. Combining them, the proposed DeepACEv2 outperforms all the previous methods, yielding the Whole Correct Ratio(WCR)(\%) with respect to images as $71.39$, and the Average Error Ratio(AER)(\%) with respect to chromosomes as about $1.17$.
\end{abstract}


\section{Introduction}
\label{sec:introduction}
Karyotyping is a cytogenetic experiment method that helps cytologists to observe the structures and features of chromosomes presented on metaphase images. In clinical practices, karyotyping generally comprises four stages: chromosome enumeration, segmentation, classification and modification, and finally, reporting results. All the above processes are based on the metaphase images of cell division generated by a microscope camera. In metaphase image, all metaphase chromosomes are stained by Giemsa staining technique to obtain G-band chromosomes where banding patterns appear alternatively darker and lighter gray-levels. Cytologists need first to pay attention to the number of chromosomes to find out numerical abnormalities that result from gaining or losing an entire chromosome. A significant proportion of abnormalities about chromosome is numerical abnormalities \cite{ref36}, which may result in some genetic diseases, such as Down syndrome \cite{ref11}. There are many business companies equipped with their microscope products with chromosome enumeration function (e.g. CytoVision \cite{ref31, ref32, ref33}, Ikaros \cite{ref34}, ASI HiBand \cite{ref35}). However, users still have to click a mouse button to label each chromosome for assisting counting in the practical process. Counting chromosomes is performed manually now on at least 20 images per patient and needs 50-100 images more when chromosome mosaicism is explored. Considering that each human cell naturally contains 46 chromosomes, it is tedious and time-consuming. Typically, a sophisticated cytologist needs 15 minutes or more to complete chromosome enumeration for one patient. Therefore, it is an urgent need to develop a computer-aided system for chromosomes enumeration.

Although some methods have been developed to solve classification \cite{ref3,ref27-cls, ref25-cls} and segmentation problems \cite{ref29-seg, ref30-seg} of chromosomes, very few of the researches have tried to establish a computer-aided method for chromosomes enumeration directly. Gajendran et al. \cite{ref1} presented a study for chromosome enumeration by combining a variety of pre-processing methods with counting algorithm, but the error rate is high. Furthermore, some segmentation methods may solve the problem indirectly, such as Arora et al. \cite{ref30-seg} and Minaee et al. \cite{ref29-seg}. However, they only focused on segmenting touching or overlapping chromosomes, and the accuracy is not high enough.


The key point of chromosome enumeration is locating and identifying each chromosome on the metaphase image accurately, which can be seen as an object detection problem. Region-based methods dominate the solution about object detection with increasingly development of deep learning method, which include two-stage methods \cite{ref8-rcnn, ref16, ref8} and one-stage methods \cite{ref18, ref19, ref22, ref37}. Compared with the latter, the former firstly need to find candidate foreground proposals, and then classify them into different classes and refine locations of them. As a result, two-stage methods are usually slower than one stage methods, but they have better performance. In this work, we employ a classical two-stage framework, Faster R-CNN\cite{ref8}, as our base framework. A powerful backbone is important to improve model performance. ResNet \cite{ref38}, as a distinguished network for image classification, has brought significant performance improvement compared to AlexNet \cite{ref39} and VGGNet \cite{ref12}, and are widely adopted as the backbone of advanced detectors. Meanwhile, Feature Pyramid Network (FPN) \cite{ref20} is proposed to attach to the backbone network for extracting and synthesizing multi-level features of the objects. Besides, researchers try to locate objects by detecting and aggregating its top left and bottom right corner points, which are tagged by unique embeddings \cite{ref37}. Furthermore, inaccurate regressions are sometimes due to that objects are severely occluded by other objects, and researchers introduced Repulsion Loss \cite{ref9} to relieve such problem and use No Maximum Suppression (NMS) \cite{ref8-rcnn} and Soft-NMS \cite{ref10} to suppress redundant predicted bounding boxes during the prediction stage. Finally, to evaluate the effectiveness of each detector, mean Average Precision (mAP) \cite{ref13} and log average missing rate ($MR^{-2}$) \cite{ref40} metrics are widely used for fair comparison.

\begin{figure}
	\centering
	\subfigure[]{\includegraphics{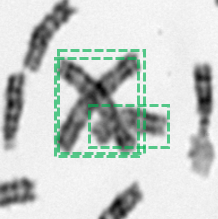}}
	\quad
	\subfigure[]{\includegraphics{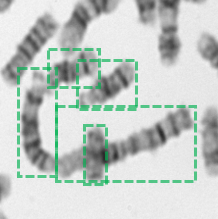}}
	\quad
	\subfigure[]{\includegraphics{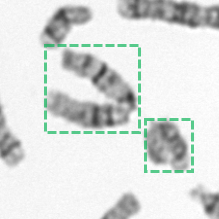}}
	\quad
	\subfigure[]{\includegraphics{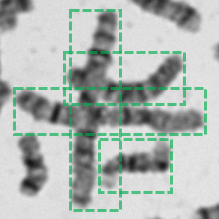}}
	\caption{The green boxes are the ground truth bounding boxes of the chromosomes. (a) shows occlusion and cross overlapping problem, two ground-truth boxes are very close to each other. (b)(c) shows the self-similarity problem. The three connected chromosomes are likely to be classified as one chromosome, and the deformed chromosomes are likely to be classified as two chromosomes connected. (d) shows a complex situation.}
	\label{fig0}
\end{figure}

The challenges for detecting chromosomes in metaphase images are mostly due to two aspects: occlusion or self-similarity.  First of all, chromosomes are floating in an oil droplet. When we generate metaphase images by a camera, chromosomes are projected to a 2-D plane which may result in severe occlusion and cross overlapping problem (Fig.\ref{fig0}(a)). The severely overlapped chromosomes may lead to inaccurate localization problems and over deletion of proposals during post-processing. Secondly, some of the chromosomes present self-similarity in G-band metaphase image (Fig.\ref{fig0}(b)): On the one hand, some partial chromosomes are similar to a whole chromosome because they have similar band patterns; On the other hand, two chromosomes are sometimes connected head to head which also brings trouble to identify, and that nonrigid chromosomes are often curved and bent which makes self-similarity problem harder to solve (Fig.\ref{fig0}(c)). The above two issues are often occurred simultaneously (Fig.\ref{fig0}(d)), which usually generate a complex chromosome cluster and make it difficult to detect all the chromosome objects accurately.

In this paper, we propose a deep learning algorithm to directly achieve chromosome enumeration on the entire G-band metaphase image, following the region-based objection detection scheme. We firstly introduce a Hard Negative Anchors Sampling (HNAS) method on Region Proposal Network (RPN) \cite{ref8} to learn more information about highly confusing partial chromosomes to solve self-similarity problem. Secondly, parallel to the detection branch, we propose a Template Module to tag each proposal by a unique 1-D embedding for heuristically separating touching and overlapping chromosomes. The embeddings generated from the Template Module are further used to guide NMS procedure to avoid over deletion of overlapped chromosomes. Furthermore, to alleviate inaccurate localization problems caused by occlusion and cross overlapping between chromosomes, inspired by the Repulsion Loss \cite{ref9}, we propose a Truncated Normalized Repulsion Loss (TNRL) and add it to detection branch where the model is jointly optimized by the combination of TNRL, classification and regression loss.A preliminary version of this manuscript was published previously on MICCAI 2019 \cite{ref43}, where a deep learning model was developed to automatically enumerate chromosomes, which significantly outperforms the model reported in \cite{ref1}. In this work, we refer to the model in \cite{ref43} as DeepACEv1. Since then, DeepACEv2, proposed in this paper, is improved with the following contributions:

\begin{itemize}
	
	\item A Hard Negative Anchors Sampling method is proposed in our previous work DeepACEv1 to solve the self-similarity problem, and we optimize the division criterion of the method in this work. We reduce the bottom threshold of hard negative anchors since background anchors, which are almost white, are easier to be correctly classified. Meanwhile, we reduce the upper threshold of hard negative anchors with a proper value to avoid conflicts of intersection over union (IoU) intervals between backgrounds of the first stage and foreground of the second stage.
	
	\item We simplify the Template Module to obtain a 1-D embedding and then combine it with the proposed Embedding-Guided NMS for heuristically separating severely overlapped chromosomes. Inspired by associative embedding mechanism \cite{ref42}, we combine distance of embeddings with IoU to determine whether two proposals belong to the same ground truth or not. To our best knowledge, it is the first attempt in chromosome studies to solve the occlusion problem by using embeddings.
	
	\item We invent a novel Truncated Normalized Repulsion Loss (TNRL) to solve the inaccurate localization problem caused by the occlusion problem. Comparing to the Repulsion Loss \cite{ref9}, TNRL is more sensitive to the object shifting from the ground truth it belongs to. It penalizes the overlapping area between the predicted bounding box and rejects ground truth but ignores the overlapping part between attracted ground truth and rejected ground truth.
	
	\item With these improvements, the mAP(\%) is increased to $71.39$, the AER(\%) is decreased to $1.17$. Besides, extensive experiments on the dataset show that improved DeepACEv2 can achieve better performance than all previous models.
\end{itemize}

The remaining parts of the paper are organized as follows: Section \ref{Method} introduces the proposed methods, including the outline of DeepACEv2 and details of each module. In Section \ref{Experiments}, we provide extensive experiments to evaluate the proposed method and discussed the benefit of each module. Finally, we conclude our work in Section \ref{Conclusion}.

\begin{figure*}
	\center
	\setlength{\belowcaptionskip}{-10pt}
	\includegraphics[width=0.9\textwidth]{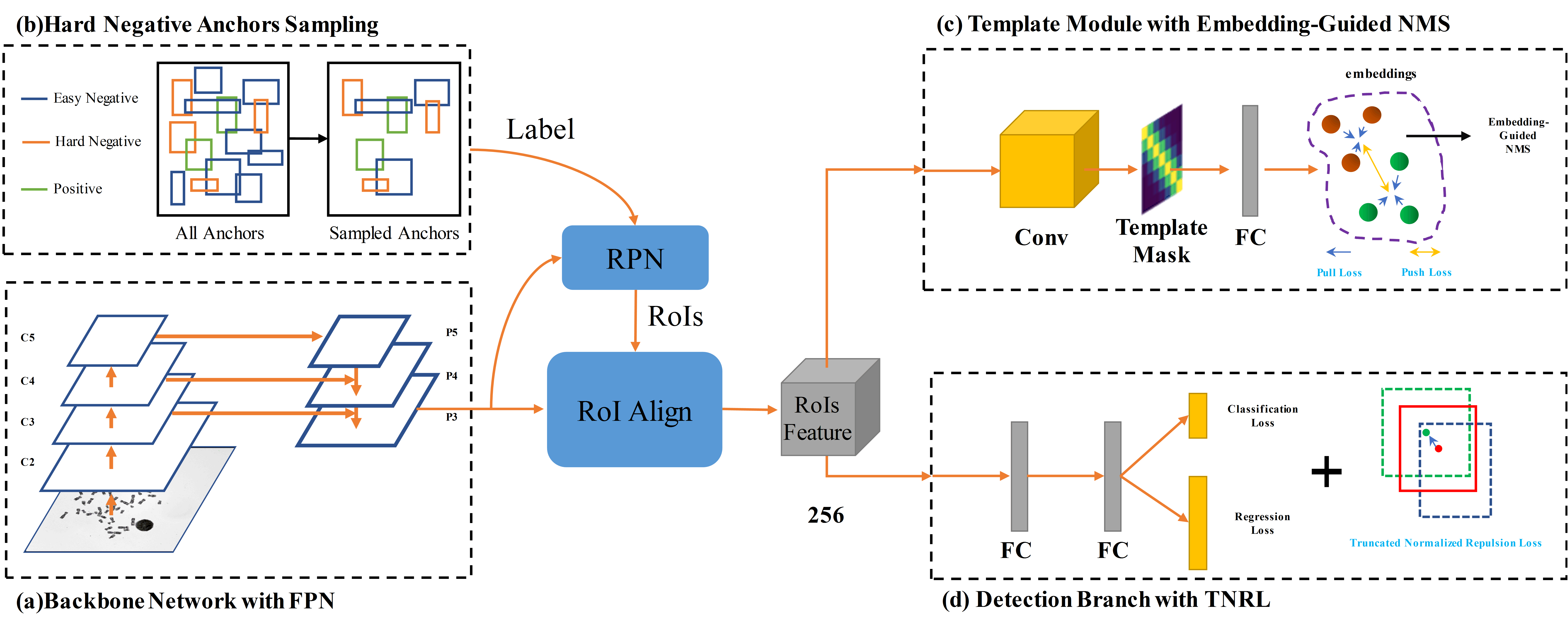}
	\caption{The framework of DeepACEv2: (a) is the picture of the Backbone Network composed of ResNet-101 and Feature Pyramid Network. (b) shows the Hard Negative Anchors Sampling procedure in the first stage. (c) illustrates the Template Module and Embedding-Guided NMS. (d) represents the classical detection branch with Truncated Normalized Repulsion Loss. The details of the Template Module are depicted in Fig.\ref{fig2}.} 
	\label{fig1}
\end{figure*}

\section{Method}\label{Method}
As shown in Fig.\ref{fig1}, besides backbone network (Section \ref{Backbone}), the proposed framework consists of three main parts: (1) candidate chromosomes detection using RPN in which a Hard Negative Anchors Sampling procedure is proposed (Section\ref{HNAS}); (2) an isolated branch with embedding based Template Module (Section \ref{Template}) and Embedding-Guided NMS (Section \ref{F-NMS}); (3) an additional Truncated Normalized Repulsion Loss (Section \ref{TNRL}). We develop the DeepACEv2  based on the region-based object detection framework \cite{ref8}. Firstly, original metaphase images collected from a clinical laboratory are taken as input to the backbone network for extracting essential features. Secondly, Region Proposal Network (RPN) take sampled anchors as training samples to classify foreground and background regions as well as refine locations. Owing to the newly designed Hard Negative Anchors Sampling method, RPN can pay more attention to the hard negative samples. Features of each candidate proposal are then cropped by classical RoIAlign \cite{ref5} method and sent into sequential fully connected layers for conventional classification and regression tasks. Parallelly, a Template Module also takes these features as input to obtain a 1-D embedding which encodes the geometric information of each candidate object. The embeddings are further used in the Embedding-Guided NMS, to alleviate missing detections caused by severe chromosomal cross overlapping during the post-processing stage. Finally, a repulsion-based loss, Truncated Normalized Repulsion Loss (TNRL), is added to the loss function to further optimize the imprecision localization caused by the occlusion.

\subsection{Backbone Network with Feature Pyramid Network}\label{Backbone}

In the original Faster R-CNN framework \cite{ref8}, a series of networks, including VGGnet and ResNet are used as the backbone network. However, those backbone networks initially designed for classifying objects with similar size (e.g. $224 \times 224$), are coarse to the localization task, especially when object sizes are small. To address this problem, Feature Pyramid Network \cite{ref20} that has a parallel "top-down" pathway to obtain feature maps with different resolutions, is proposed, which can extract features from proper resolution according to object sizes. In this work, we build our model with the ResNet-101 backbone and connect it to the FPN. As shown in Fig.\ref{fig1}(a), ResNet-101 $(C3, C4, C5)$ is used to extract high-level information of objects, and FPN $(P5, P4, P3)$ is used to enlarge the resolution of feature maps and combine high-level information with low-level information. It is worth noting that most sizes of chromosomes are between $32$ pixel and $128$ pixel. Therefore, it is enough to cover the sizes of chromosomes in the dataset by only choosing the $P3$ as the output level.

\begin{figure}
	\centerline{\includegraphics[width=\columnwidth]{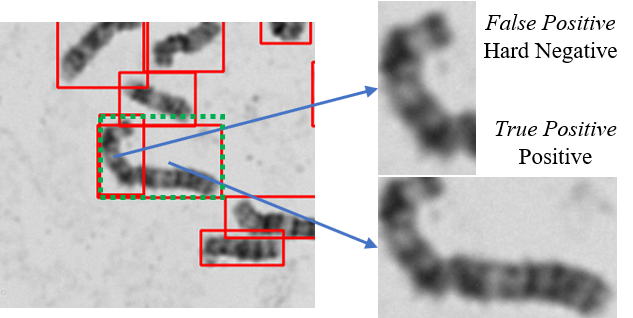}}
	\caption{An example of the similarity between whole chromosomes and partial chromosomes, green and red boxes represent ground truths and predicted bounding boxes, respectively.}
	\label{fig3}
\end{figure}

\subsection{Hard Negative Anchors Sampling in Region Proposal Network}\label{HNAS}

The region-based object detection models, such as Faster R-CNN \cite{ref8}, firstly introduce a region proposal network (RPN) to generate candidate proposals. Typically, RPN only focuses on the binary classification of integrated objects (eg. IoU $\geq 0.7$) and background (eg. IoU $<0.3$). The selected proposals are passed to Fast R-CNN for further fine classification and regression in which partial objects (eg. $0.3\leq$ IoU $<0.7$) are taken care of. However, unlike natural object, as shown in Fig.\ref{fig3}, chromosomes usually have various length and similar banding patterns, which confuses the network to discriminate partial and whole chromosomes, namely self-similarity problem. Meanwhile, the risk of irreversibly losing information in Fast R-CNN, such as cropping features and RoI Pooling (or RoIAlign, notice that in our framework, we used RoIAlign instead of the original RoI Pooling method to crop feature), also makes the network hard to distinguish partial chromosomes. To this end, we propose a novel Hard Negative Anchors Sampling method during the RPN sampling procedure to better identify partial chromosomes and solve the self-similarity problem. 

In our previously published DeepACEv1, we define those anchors that have an IoU in the $[0.3,0.7)$ as hard negative anchors for clarity and the original negative anchors with an IoU $<0.3$ are named as easy negative anchors. In this work, we improve the partition criterion by considering the properties of data and the Faster R-CNN model simultaneously. First of all, as Fast R-CNN assigns positive and negative labels to candidate proposals based on an IoU of $0.5$, to avoid feature semantic conflict of RPN and Fast R-CNN in the interval $[0.5, 0.7)$, we only apply Hard Negative Anchors Sampling on anchors that have the IoU lower than $0.5$. Additionally, we reduce the bottom threshold of hard negative anchors to $0.1$ because background anchors (IoU$\le 0.1$), which are almost empty, are easier to be correctly classified. Therefore, we regard anchors that have IoU in the interval $[0.1, 0.5)$ as hard negative anchors to make the partial chromosomes trained by the RPN more sufficiently.

Considering that RPN suffers from severe inter-class imbalance (positive $:$ negative $\approx1:4000$) and intra-class imbalance (hard negative $:$ easy negative $\approx1:4$), a new Hard Negative Anchors Sampling method inspired by stratified sampling is then proposed. As shown in Fig.\ref{fig1}(b), we divide all anchors into the positive, hard negative, and easy negative anchors according to IoU overlap with ground truth box. We use mini-batches of size R=512 for training RPN and take 25\% of the anchors from positive. Half of the remaining are uniformly sampled from hard negatives (37.5\%), and the rest are sampled from easy negatives (37.5\%). Finally, positive anchors are labeled with a foreground object class, both hard and easy negative anchors are labeled as background; the loss function is the same as the original RPN. In this way, feature maps generated by RPN are enhanced by hard negative anchors information, and the following stage is improved by these features.

\subsection{Template Module for Disentangling Occlusion Chromosomes}\label{Template}
\begin{figure}[H]
	\centerline{\includegraphics[width=\columnwidth]{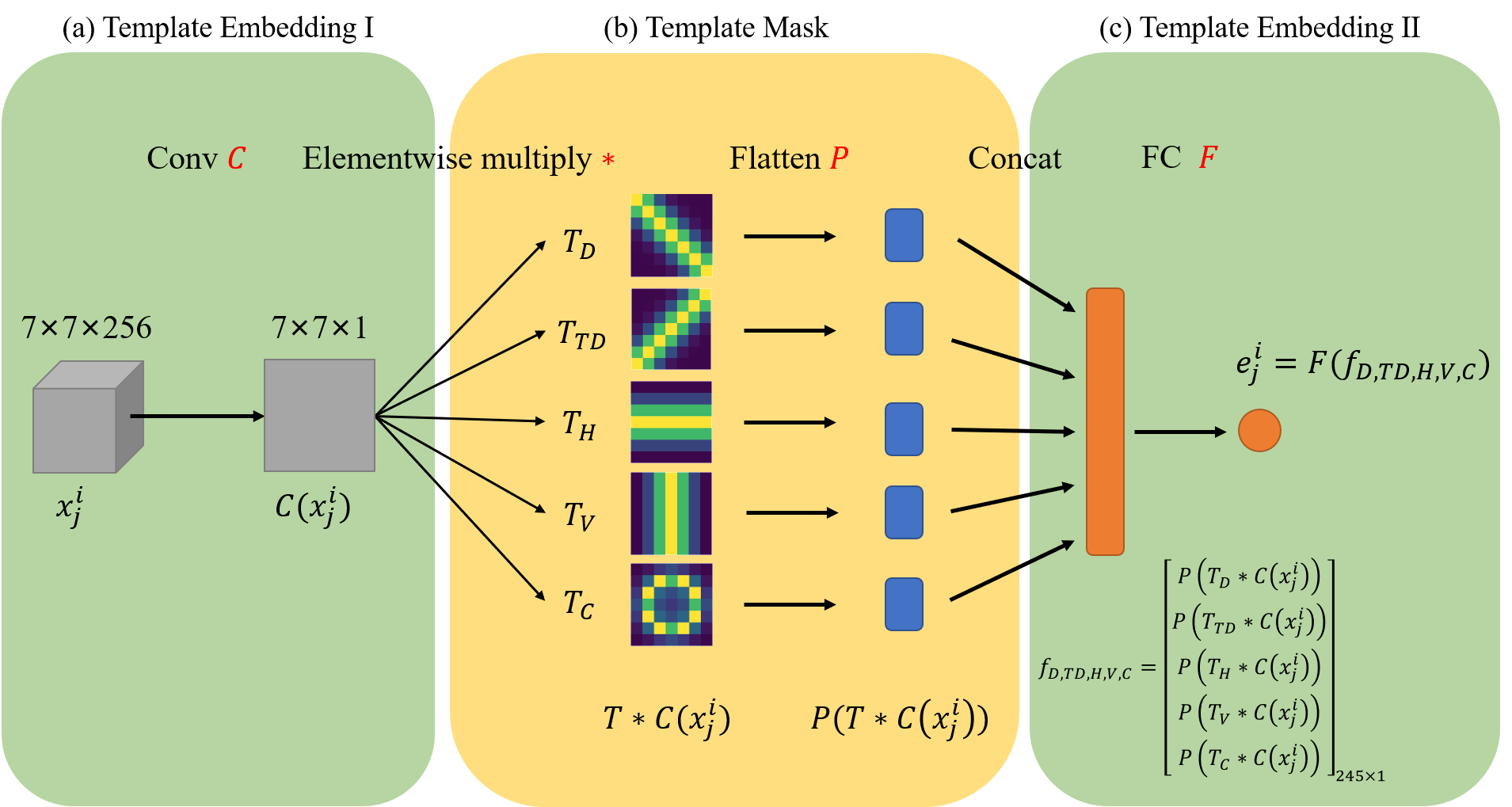}}
	\caption{The illustration of the Template Module branch of DeepACE: (a) part \uppercase\expandafter{\romannumeral1} of Template Embedding Block is a convolutional layer with ReLU. (b) shows the Template Mask, which extracts the specific features of the proposal and flattens them to 49-d. (c) par \uppercase\expandafter{\romannumeral2} of Template Embedding Block, which is a fully connected layer.}
	\label{fig2}
\end{figure}

Touching and overlapping chromosomes bring severe intra-class occlusion, in which a network cannot localize and identify each chromosome correctly. To alleviate this problem, we add an embedded Template Module as an individual branch to heuristically separate the touching or overlapping chromosomes. Specifically, although chromosomes are usually displayed with bending or deformation in metaphase images, they can be summarized into some regular schemes. Therefore, it is reasonable to introduce several general-template masks to represent patterns of chromosomes. When two or more chromosomes are overlapped together within a selected proposal, a particular chromosome can be extracted by the corresponding template mask, and thus facilitates the separation of overlapping chromosomes. We summarize the implementation details of the template module in Fig.\ref{fig2}.

The design of template masks is essential to influence performance. However, features will be affected by extra ground truth if we directly take ground truth of overlapped chromosomes as template masks. Regarding the geometric characteristic of chromosomes, we observe that chromosomes are usually displayed as slender strips in labeled bounding boxes and locates along the diagonal or horizontal or vertical direction. As a result, peak values are located on the central part of the feature map along the diagonal or horizontal or vertical direction, which leads to $T_D$, $T_{TD}$, $T_{H}$ and $T_{V}$. Besides, a circle-like template mask $T_{C}$ is introduced since there are a few seriously bending chromosomes, as shown in Fig.\ref{fig5}(c). The feature map size is $7{\times}7$, we introduce $ID_{row}\in \{0, 1,2,3,4,5,6\}$ and $ID_{col}\in \{0, 1,2,3,4,5,6\}$ to indicate the pixel's location of feature map, all the five template masks are designed as constant matrix with Gaussian distribution, where $x_{row} = ID_{row} -3 \ , \ y_{col}=ID_{col}-3$:

\begin{equation}
	\small
	\begin{aligned}
		T_D(ID_{row}, ID_{col}) &= e^{-\frac{(x_{row}-y_{col})^2}{3}} \\
		T_{TD}(ID_{row}, ID_{col}) &= e^{-\frac{(x_{row}+y_{col})^2}{3}} \\
		T_H(ID_{row}, ID_{col}) &= e^{-\frac{y_{col}^2}{3}} \\
		T_V(ID_{row}, ID_{col}) &= e^{-\frac{x_{row}^2}{3}}\\
		T_C(ID_{row}, ID_{col}) &=e^{-\frac{|x_{row}^2+y_{col}^2 - 5|}{3}}
	\end{aligned}
\end{equation}

After RoIAlign, feature maps with the shape of $7 \times 7 \times 256$ are separately sent into a template module and a detection pathway composed of two fully connected layers. Same as the original Faster R-CNN, the detection pathway returns a binary classification score and a regressed bounding box. The template module pathway is used to obtain the embedding of each candidate proposal, to further determine whether a pair of proposals are from the same ground truth bounding box. Inspired by the Associative Embedding method proposed in \cite{ref37,ref42}, we group candidate proposals according to the distance of their embeddings. As shown in Fig.\ref{fig2}, the template module is composed of a template embedding block and a template mask block, in which the template mask block is located in the middle of the template embedding block. Specifically, the first part of the template embedding block (as shown in Fig.\ref{fig2}(a)) fuses the features of each proposal along the $256$ channels using a $1 \times 1 \times 256$ convolutional layer. Next, based on the five template masks, the template mask block (as shown in Fig.\ref{fig2}(b)) extracts features of specific locations followed by the flattening operation. Subsequently, the five flattened features are concatenated into one 245-D ($7 \times 7 \times 5$-D) vector. Finally, as pointed in \cite{ref42}, a 1-D embedding is sufficient for multiple pattern estimation, we (Fig. \ref{fig2}(c)) apply a $245 \times 1$ fully connected layer to encode the 245-D feature into a 1-D embedding output. The 1-D embeddings will play an essential role in the post-processing procedure, as detailed in Section \ref{F-NMS}.

Following the principle of associative embedding method, the distance of two proposals belong to the same ground truth should be minimized. On the contrary, if two proposals belong to different ground truth, we should maximize their distance. Following the strategies used in \cite{ref37,ref42}, we design a grouping loss which is linear weighted sum of pull loss and push loss ($\alpha L_{pull} +\beta L_{push}$, $\alpha, \beta$ are weight parameters). The pull loss $L_{pull}$ is used to minimize the distance between embeddings that belong to the same ground truth and can be defined as:
\begin{equation}
	L_{pull} = \frac{1}{N}\sum_{j=1}^{N_{gt}}\sum_{i=1}^{N_{j}^p}(e_j^i-\bar{e}_j)^2
\end{equation}
Where $N_{gt}$ means the total number of ground truth and $N_j^p$ is the number of proposals that belong to $j$-th ground truth. $e_j^i$ is the embedding for the $i$-th proposal of the $j$-th ground truth and $\bar{e}_j= \frac{1}{N_{j}^p}\sum_{i=1}^{N_{j}^p}e_j^i$ represents the mean value of all the embeddings of the proposals belong to  $j$-th ground truth. $N$ is the total number of positive proposals. In the following, we define $\hat{e}_j^i=|e_j^i-\bar{e}_j|$ for notational convenience. 

It has been discovered in RetinaNet \cite{ref44} that a large number of samples are easy to be optimized. As the training goes on, the contribution of the loss value on each sample is small, but the summation of them can be large and may even dominate the loss term. Therefore, similar to RetinaNet, We design a soft weighting factor $(\theta + \hat{e}_j^i)^\lambda$ to make the network focus more on the hard samples during training. The modified pull loss is shown as: 
\begin{equation}\label{pull_loss}
	L_{pull}=\frac{1}{N}\sum_{j=1}^{N_{gt}}\sum_{i=1}^{N_{j}^p}[(\theta+ \hat{e}_j^i)^\lambda \cdot \hat{e}_j^i{}^2]
\end{equation}
where $\theta\ge 0$ is a threshold value to divide proposals into easy samples and hard samples. Similar to that in the RetinaNet, $\lambda \ge 0$ is a tunable focusing parameter. In this way, the samples $\hat{e}_j^i > 1 - \theta$ will be treated as hard samples and have greater effects on the pull loss. The pull loss $L_{pull}$ is only applied to the positive candidate proposals during training.

Similar to \cite{ref37,ref42}, we also employ a push loss $L_{push}$ to provide a penalty when the distance between embeddings of two different ground truth are smaller than a given threshold. Considering that the embeddings are mainly used to separate two severely overlapped proposals, we only apply the push loss on the ground truth which has the highest IoU with a given ground truth. Given a ground truth $g_i \in \mathcal{G}$, $g_i^{rep}$ is defined as the ground truth that has the highest IoU with $g_i$ except itself, called repulsion ground truth of $g_i$:
\begin{equation}
	g_i^{rep}=\mathop{\arg\max}_{g\in \mathcal{G}\verb|\|\{g_i\}}IoU(g_i, g)
\end{equation}
All the isolated ground truths are not considered here, the push loss  $L_{push}$ is defined as:
\begin{equation}
	L_{push}=\frac{1}{N_{gt}'}\sum_{i=1}^{N_{gt}'} \max(\delta-|\bar{e}_{g_i}-\bar{e}_{g_i^{rep}}|, 0)
\end{equation}
where $N_{gt}'$ is the total number of ground truths which are not isolated and $\delta>0$ is the distance threshold, and we set $\delta$ to be $1$ in all our experiments. Both $\bar{e}_{g_i}$ and $\bar{e}_{g_i^{rep}}$ are the mean values of the embeddings of the corresponding ground truth. 

\subsection{Embedding-Guided NMS}\label{F-NMS}
In post-processing stage, IoU based algorithms like Non-Maximum Suppression (NMS) \cite{ref8-rcnn} and Soft-NMS \cite{ref10} are widely used in recent years. They suppress redundancies according to the IoU metric, in which highly overlapped predicted bounding boxes are removed directly or inhibited through decaying its detection scores. However, over deletion still frequently happens when severe occlusion occurs. Thus, we propose the Embedding-Guided NMS based on Soft-NMS, which introduces embedding of proposals to optimize the score decay function. The basic idea is that if embeddings of two bounding boxes are far away, they should represent two different chromosomes. Therefore, in Embedding-Guided NMS, we compute the distance $d$ between embeddings and assign a threshold value of $\Delta$ (set as $0.3$). We will lighter decay score if $d> \Delta$ and heavier decay score if $d<\Delta$. The overview of the algorithm is summarized in Algorithm \ref{alg:fg_nms}.
\begin{algorithm}[htb] 
	\begin{small}
		\caption{Feature-Guided Non Maximum Suppression} 
		\label{alg:fg_nms} 
		\begin{algorithmic}[1] 
			\REQUIRE ~~\\ 
			The list of initial detection boxes $B=\{b_1, \dots, b_N\}$;\\ 
			The list of corresponding detection scores $S=\{s_1, \dots, s_N\}$;\\
			The list of corresponding embeddings $E=\{e_1, \dots, e_N\}$;
			\ENSURE ~~\\ 
			The list of detection boxes with new order $D'$;\\
			The list of corresponding detection scores which are decayed by function $S'$
			\STATE Initialize $D'=\{\}$; 
			\WHILE{$B \neq \{\}$} 
			\STATE Sort all the detection boxes $B$ by scores $S$ in descending order, mark the first candidate as $b_{max}$, corresponding score $s_{max}$ and embedding $e_{max}$
			\STATE Append the $b_{max}$ into $D'$ and pop it from $B$
			\STATE Append the $s_{max}$ into $S'$ and pop it from $S$
			\FOR{$b_i \in B$}
			\STATE Measure the distance $d=|e_{max}-e_i|$
			\STATE Sigmoid decay $\mathop{S}(d)=\frac{1}{1+e^{-2(d-\Delta)}}$
			\STATE Compute new score $s_i$: $s_i=s_ie^{-\frac{iou(b_{max}, b_i)^{(1.5+\mathop{S}(d))}}{\sigma}}$
			\ENDFOR
			\ENDWHILE
			\RETURN $D'$ and $S'$
		\end{algorithmic}
	\end{small}
\end{algorithm}

\vspace{-10pt}
\subsection{Truncated Normalized Repulsion Loss}\label{TNRL}
We further propose a new loss function aiming at alleviating the influence caused by severe intra-class occlusion in karyotyping images. Specifically, the new repulsion loss function $L_{TNRep}$ is proposed to prevent predicted boxes from shifting to adjacent objects when occlusion of ground truths occurred.

Assume \( \mathbb{P}_{+}\) is the set of positive proposals produced by RPN network and \(P \in \mathbb{P}_+\), $B^P$ is the predicted box of $P$, $G^P$ is the designated target of $P$ defined as the ground truth that has the highest IoU with $P$, $R^P$ is the repulsion ground truth object of $P$ defined as the ground truth that has the highest IoU with $P$ except $G^P$. The repulsion loss term is firstly proposed in Wang et al. \cite{ref9} as:

\begin{equation}\label{equa2}
	L_{Rep} = \frac{\sum_{P \in \mathbb{P}_+} {\text{Smooth}_{ln}}(IOG(B^P,R^P))}{\vert \mathbb{P}_+ \vert}
\end{equation}

Where $IOG(A,B) \triangleq \frac{area(A \cap B)}{area(B)}$, and $\text{Smooth}_{ln}$ is a smooth function.

\begin{figure}
	\centering
	\subfigure[]{\includegraphics[width=\columnwidth]{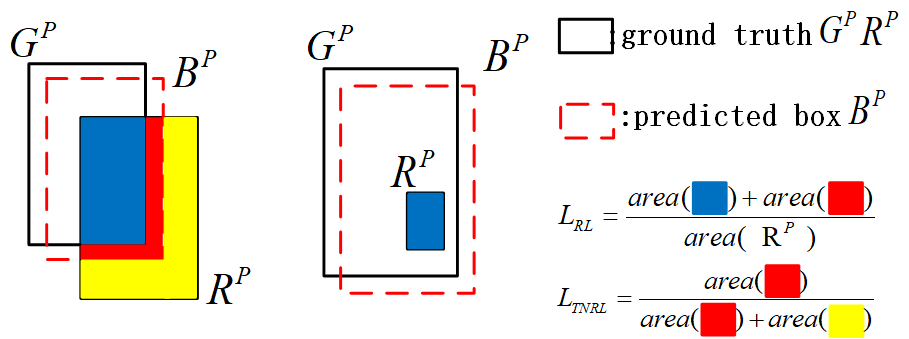}}
	\quad
	\subfigure[]{\includegraphics[width=\columnwidth]{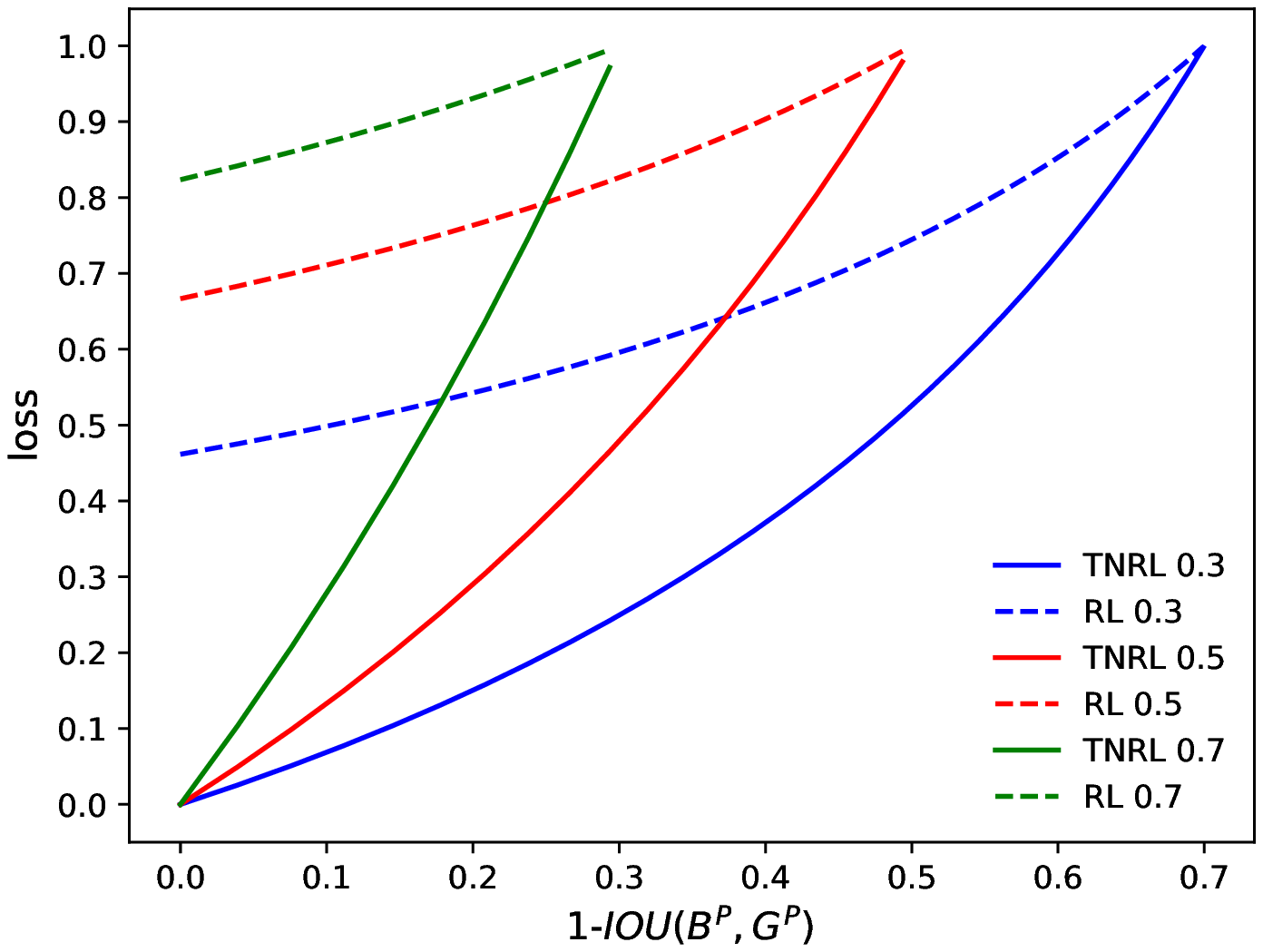}}
	\caption{(a) demonstrates the difference between TNRL and RL. (b) is the comparison of TNRL and RL with different shifting errors, at different occlusion situations($IOU(G^P,R^P)$ values are set as $0.3$, $0.5$, $0.7$).}
	\label{fig4}
\end{figure}

However, as shown in the first illustration of Fig.\ref{fig4}(a), if the overlap of two ground truth bounding boxes $G^P$ and $R^P$ is severe, the huge intersection part will dominate the repulsion loss even when $B^P$ matches $G^P$ well. In this situation, the repulsion loss will not be sensitive to shifting. Also, the repulsion loss will always be equal to 1 if the ground truth target of $B^P$ contains a small ground truth target, as shown in the second illustration of Fig.\ref{fig4}(a). These phenomena limit the ability of the repulsion loss function to localize each chromosome accurately. Furthermore, Faster R-CNN may be hard to converge when $G^P$ and $R^P$ are overlapped heavily because of the large loss value. Therefore we propose a novel repulsion loss function, called Truncated Normalized Repulsion Loss as:

\begin{equation}
	\begin{aligned}
		&IOG'(B^P, R^P, G^P)=\frac{IOG(B^P,R^P)-IOG(G^P,R^P)}{1-IOG(G^P,R^P)} \\
		&L_{TNRep} = \frac{\sum_{P \in \mathbb{P}_+} \text{Smooth}_{ln}(\max(IOG'(B^P, R^P, G^P),0))}{\vert \mathbb{P}_+ \vert}
	\end{aligned}\label{equa3}
\end{equation}

Where,

\begin{equation}\label{equa4}
	\begin{split}
		IOG'(B^P, R^P, G^P) &=\frac{area(B^P\cap R^P)-area(G^P \cap R^P)}{area(R^P)-area(G^P \cap R^P)}
	\end{split}
\end{equation}

The comparison of Truncated Normalized Repulsion Loss (TNRL) and original Repulsion Loss (RL) is depicted in the right equations of Fig.\ref{fig4}(a). Similar to the original repulsion loss, the novel loss function can only be decreased by decreasing $area(B^P \cap R^P)$. But the loss depends on the shifting of the prediction box only. It is not affected by the overlapping of $G^P$ and $R^P$, which means the loss value changes more directly according to the severity of shifting. Meanwhile, it has an upper bound equal to one if $B^P$ is coincident with $R^P$; and a lower bound equal to zero when it's coincident with $G^P$, which means the range of the loss value is greater than the original repulsion loss depending on the severity of the prediction error, especially in the severe occlusion situation Fig.\ref{fig4}(b\footnote{For simplify, the diagram is depicted when two ground truths are equal in size, and the predicted box is shifted from one to another straightly.}).Furthermore, when $G^P$ contains $R^P$, the new loss is equal to zero rather than confused by a large value when applying the original repulsion loss. Therefore, our new loss function is a more accurate representation of the shifting error (as shown in Fig.\ref{fig4}(b)). The experiments in Section \ref{Ablation} proved that the novel loss function regresses proposals more precisely.

\vspace{-10pt}
\section{Experiments}\label{Experiments}
\subsection{Datasets}
To validate the proposed method on the entire metaphase image, we collect $1375$ Giemsa stained microscopic metaphase images containing $63026$ objects from the Peking University Third Hospital. All grayscale metaphase images come from Leica's CytoVision System (GSL-120) with a resolution of $1600 \times 1200$. All images are labeled by a cytologist with a rectangle bounding box associate with each chromosome and then verified by another cytologist. We randomly split images into $3:1:1$ as training set($825$), validation set($275$) and testing set($275$) and combine training set and validation set as trainval set. All images in the training set are used for training, and the validation set is used for ablation study and hyper-parameter searching. We use the trainval set to train different methods and report final results on the testing set for fair comparisons. All the above information about datasets are shown in Table \ref{data}.

\begin{table}[H]
	\caption{Details of our datasets.}
	\label{data}
	\centering
	\begin{tabular}{ccccc}
		\toprule[2pt]
		Datasets & Image \# & Object \# & Resolution & Total \# \\
		\midrule[1pt]
		training & 825 & 37819 & $1600 \times 1200$ & \multirow{3}{*}{1375} \\
		validation & 275 & 12593 & $1600 \times 1200$ & \\
		testing & 275 & 12614 & $1600 \times 1200$ & \\
		\bottomrule[2pt]
	\end{tabular}
\end{table}

\subsection{Evaluation Metrics}
To abundantly evaluate the performance of DeepACEv2, we introduce six metrics: Whole Correct Ratio (WCR), Average Error Ratio (AER), $F_1$-score, Accuracy (Acc), Mean Average Precision (mAP) and Log Average Missing Rate ($MR^{-2}$). To measure the overall performance of models, we choose mAP and $MR^{-2}$ as evaluation metrics, which are designed for evaluating object detection methods and pedestrian detection methods, individually. Except for the mentioned metrics, WCR, AER, $F_1$-score, and Acc are used to measure the model performance under a given condition (for example, detection confidence of 0.5). Besides, comparing to the $F_1$-score and Acc, WCR and AER have more clinical meanings.

Both mAP and $MR^{-2}$ as traditional evaluation metrics have been defined in \cite{ref13} and \cite{ref40}, in which a higher mAP is better and it is contrary for $MR^{-2}$. Specifically, methods output a bounding box with a confidence score for each detection. Next, we decide whether a prediction is correct or not according to the following three basic criteria. All predictions need to be ranked by the decreasing confidence of the "chromosome" class and then used to compute the precision-recall curve on the "chromosome" class. Finally, the mAP is the area under the curve. Besides, in some certain tasks such as chromosome enumeration, $MR^{-2}$ is preferred to mAP since there is an upper clinical limitation on the acceptable false positives per images (FPPI) rate. We compute $MR^{-2}$ by averaging miss rate at nine FPPI rates evenly spaced in log-space in the range $10^{-2}$ to $10^{0}$.

Before computing the remaining four metrics, we firstly define the following three basic criteria:
\begin{itemize}
	\item True Positive ($TP_K$): The predicted bounding box is a true positive if it satisfies following two conditions: (a) it can be assigned to a ground truth because IoU of this pair is highest among all ground truths and above a given threshold (0.5 in this study) at the same time; (b) it has the highest score among all the proposals that assigned to this ground truth. $TP_k$ is the total number of true positives of $k$-th images.
	\item False Positive ($FP_k$): The predicted bounding box that does not have an IoU greater than a threshold with any ground truth or has the max IoU with a ground truth that has already been detected is a false positive. $FP_k$ is the total number of false positives of $k$-th images.
	\item False Negative ($FN_k$):The ground truth that is not detected by any predicted bounding box is a false negative. $FN_k$ is the total number of false negatives of $k$-th images.
\end{itemize}
In all experiments, we set the threshold of $0.5$ to define whether the predicted bounding box is true positive or not.

The $F_1$-score is computed as:
\begin{equation}
	\begin{aligned}
		F_1 =\frac{2\cdot Precision \cdot Recall}{Precision + Recall}, \\
		Precision = \frac{\sum_{k}TP_k}{\sum_{k}(TP_k+FP_k)}, \\
		Recall = \frac{\sum_{k}TP_k}{\sum_{k}(TP_k+FN_k)}
	\end{aligned}
\end{equation}
Meanwhile, the miss rate used in $MR^{-2}$ can be computed by the value of $1-Recall$ under the threshold determined by FPPI rate.

The accuracy (Acc) is adopted to measure whether positive and negative proposals are classified correctly. Because all outputs of the model are proposals classified as the ”chromosome” class which are either True Positives (TP) or False Positives (FP), we do not include the true negative (TN) term in Acc:
\begin{equation}
	\text{Acc} = \frac{\sum_{k}TP_k}{\sum_{k}(TP_k+FP_k+FN_k)}
\end{equation}

The AER is defined as the fraction of the sum of false positives and false negatives divided by the number of ground truth:
\begin{equation}
	\text{AER}=\frac{\sum_k(FP_k+FN_k)}{\sum_k(TP_k+FN_k)}
\end{equation}

The WCR is defined as the percentage of all right images in the whole testing set. Assume $N_+$ is the number of images where all ground truths have been properly detected ($FN_j=0$), and no false positives remained ($FP_j=0$), and $N$ is the number of total images in validation set or the testing set. Then, WCR can be computed by $N_+ / N$. Different from other metrics that evaluate performance on instance-level, WCR is used to evaluate the performance of the model on the image-level. Only when all the chromosomes of an image are correctly detected, the WCR improve. Therefore, it is a stricter and more sensitive criterion after the model reaching a decent level of discriminability.

\subsection{Implementation Details}
\begin{figure}
	\centering
	\subfigure[]{\includegraphics[width=0.45\columnwidth]{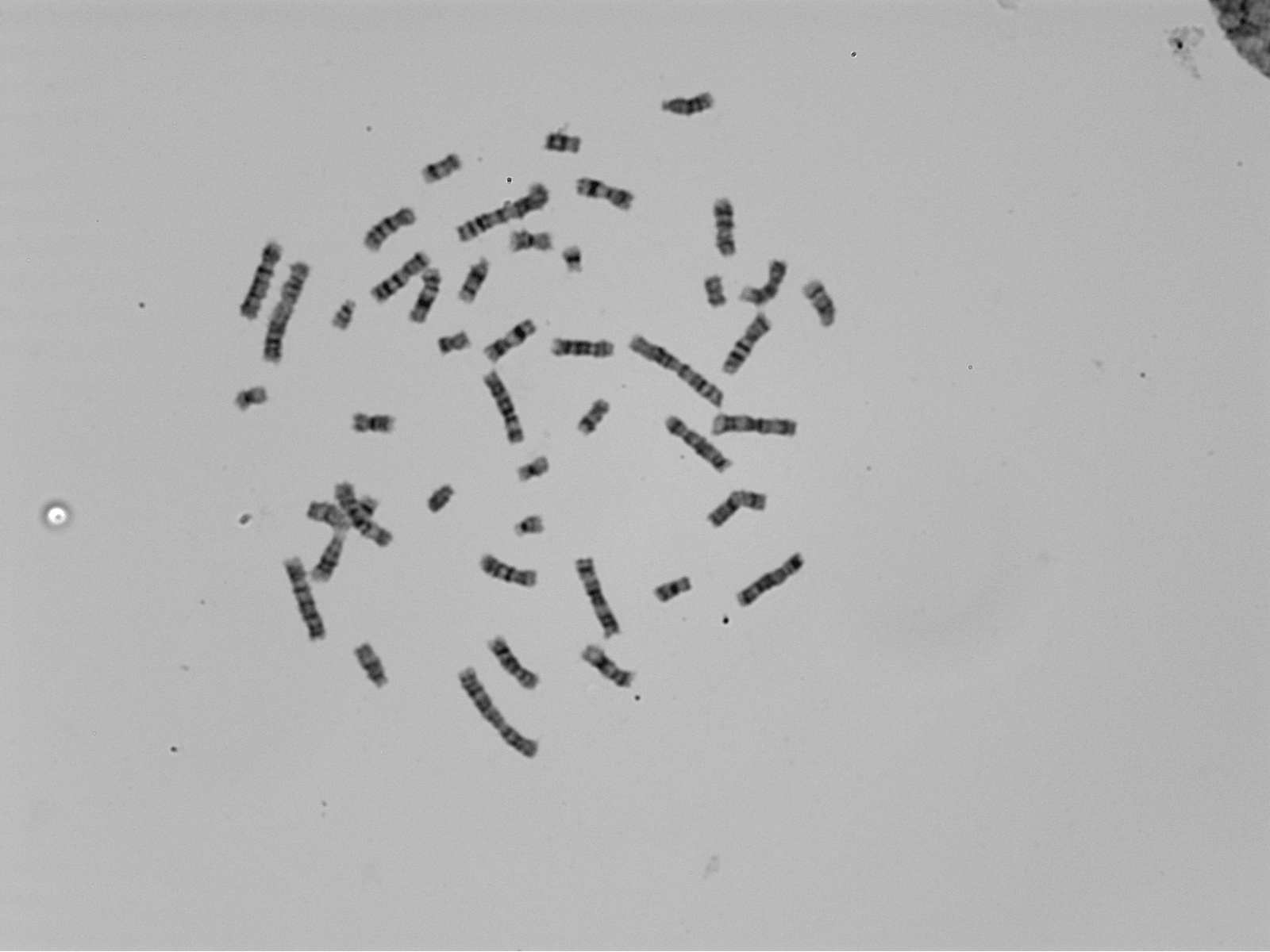}}
	\subfigure[]{\includegraphics[width=0.45\columnwidth]{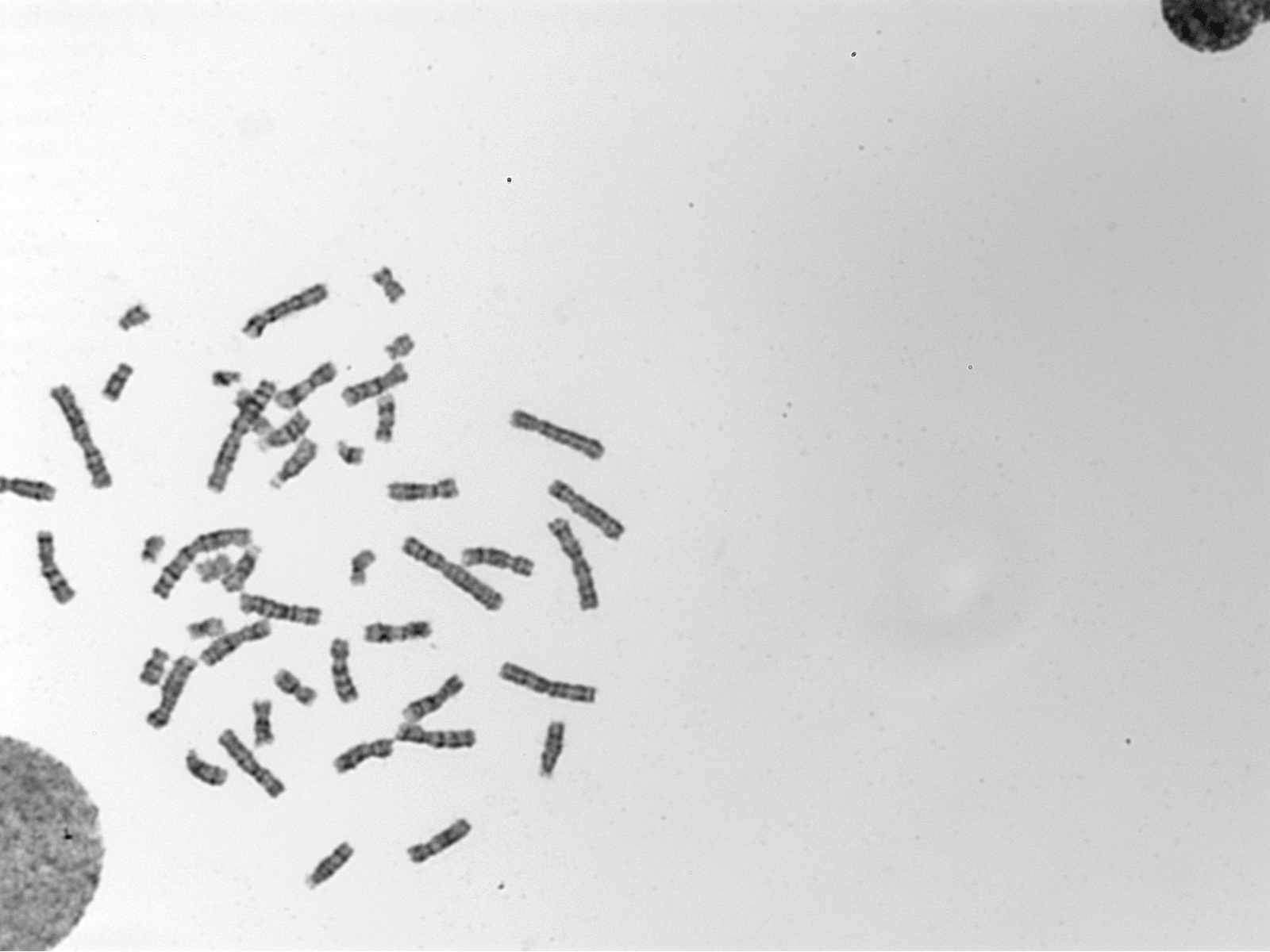}}
	\subfigure[]{\includegraphics[width=0.45\columnwidth]{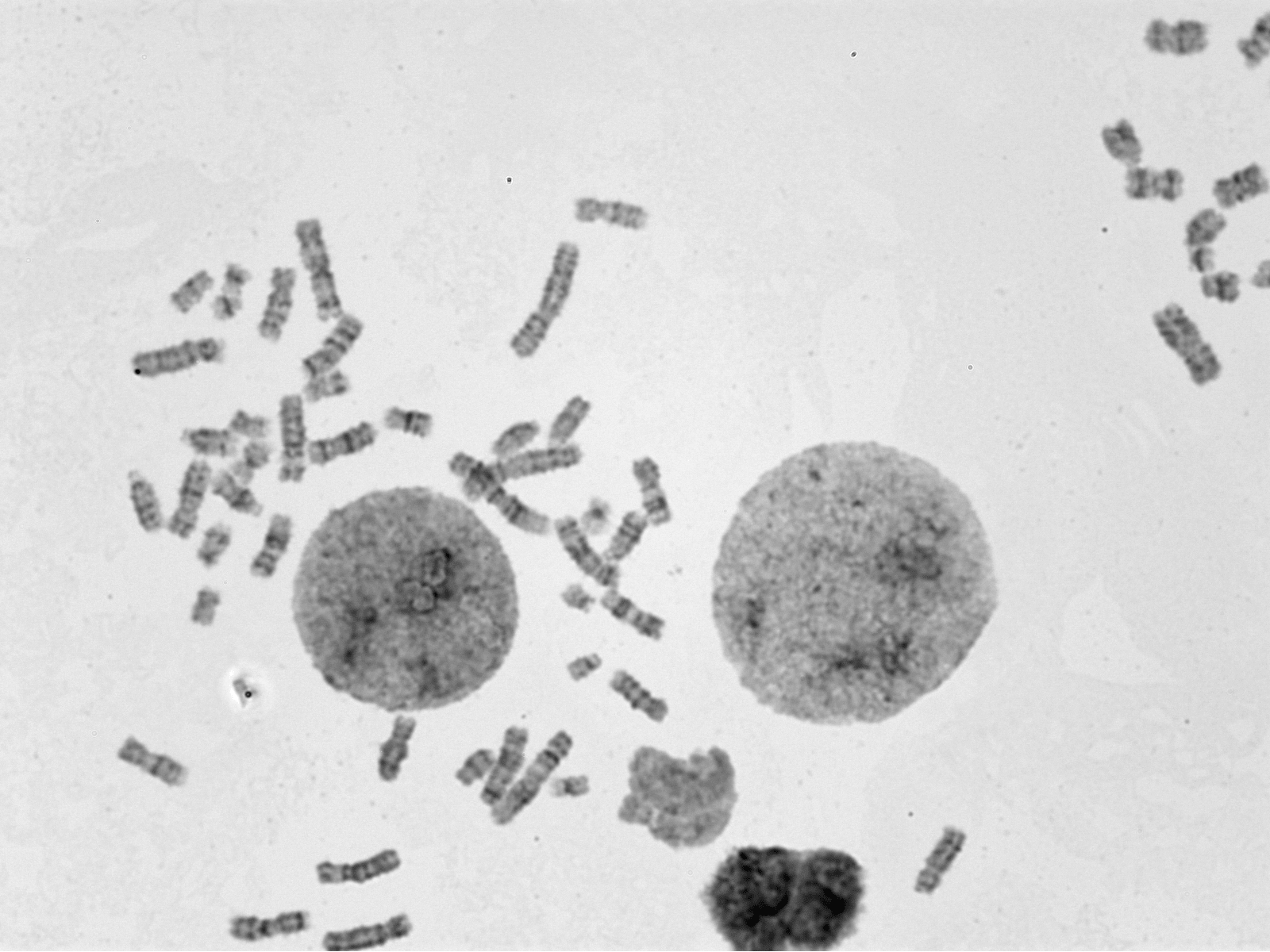}}
	\subfigure[]{\includegraphics[width=0.45\columnwidth]{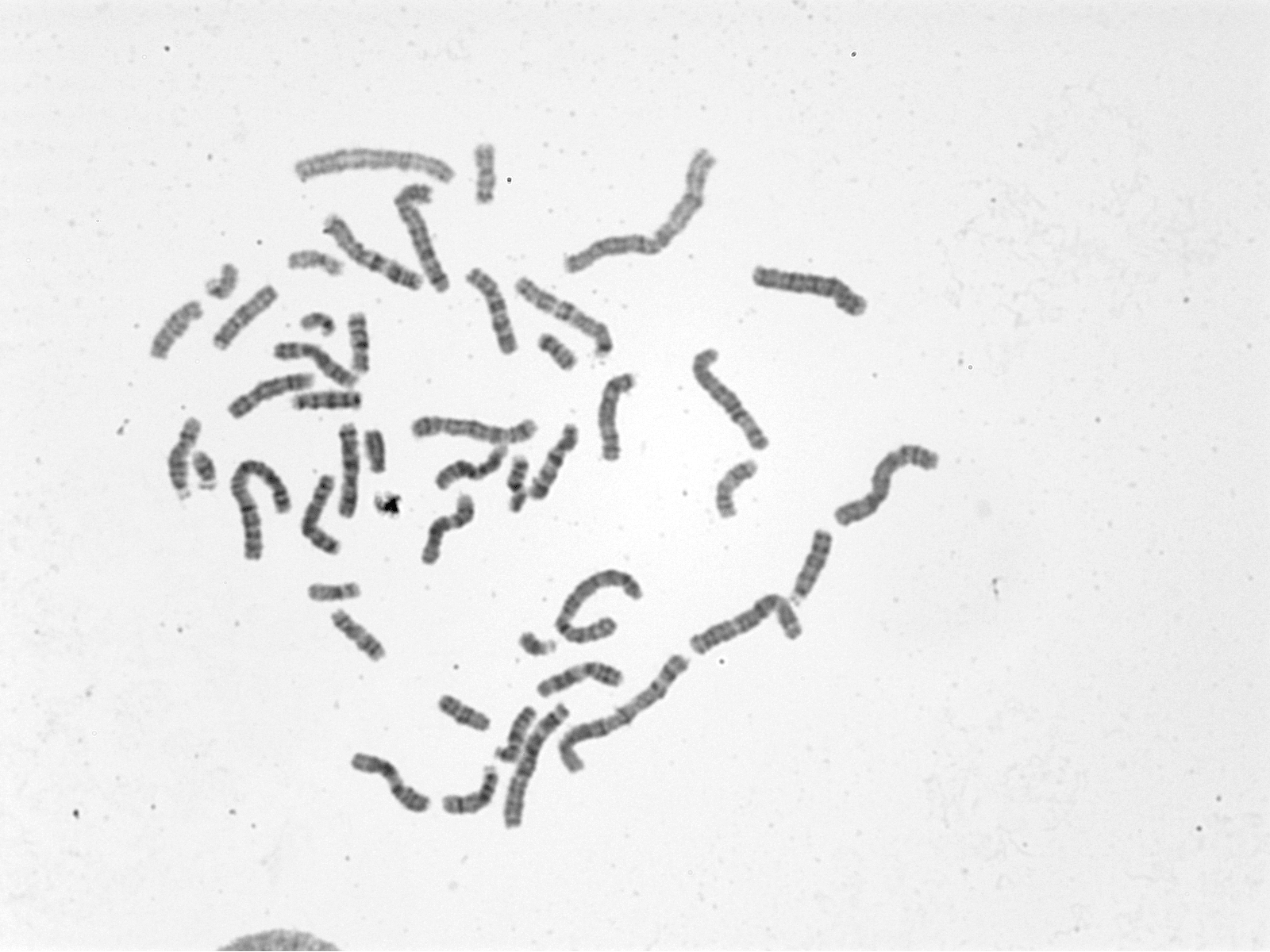}}
	\caption{Giemsa stained metaphase images exported from microscopes.}
	\label{img_example}
\end{figure}

The network is implemented on the MMDetection toolbox based on PyTorch \cite{ref45} deep learning library. The backbone and detection branches are initialized under the default setting of the MMDetection toolbox, where the backbone is pre-trained on the ImageNet dataset \cite{ref46}. In Template Module, Conv layer and FC layer is initialized by He initialization \cite{ref47}. Giemsa stained metaphase images exported from microscopes suffer from diversity impurity and contrast, as shown in Fig. \ref{img_example}. However, benefit from the strong representation power of deep learning, it is enough to normalize images by the mean value and standard deviation in the pre-processing stage.

During training, we conduct random horizontal flipping as data augmentation for reducing overfitting. A flipping operation will apply on each image with a probability of $0.5$ to horizontally flip the whole image as well as all its ground truth bounding boxes. We set anchors that have areas of $\{32^2, 64^2, 96^2, 128^2\}$ with $9$ aspect ratios $\{1\text{:}5, 1\text{:}4, 1\text{:}3, 1\text{:}2, 1\text{:}1, 2\text{:}1, 3\text{:}1, 4\text{:}1, 5\text{:}1\}$. After RPN, according to hyperparameter selection, we choose the top $6000$ proposals to apply NMS and remain at most $2000$ proposals to Fast R-CNN. Both RPN and Fast R-CNN modules are jointly end-to-end optimized during training. We use Stochastic Gradient Descent (SGD) to optimize the full training loss:
\begin{equation}
	L=L_{det} + \alpha L_{pull} + \beta L_{push} + \gamma L_{TNRep}
\end{equation}
where $\alpha$, $\beta$ and $\gamma$ are the weights for the pull, push, and truncated normalized repulsion loss, respectively. $L_{det}$ is the original losses of Faster R-CNN, including classification and regression losses of RPN and detection head. We set both $\alpha$ and $\beta$ to $0.1$ and $\gamma$ to $0.5$. We set $\theta$ and $\lambda$ of Eq. \ref{pull_loss} in Template Module to $0.5$ and $2$ individually. We train the model for a total of $24$ epochs with a mini-batch of $2$ images. The initial learning rate of $0.02$, and it is decreased by $0.1$ after $16$ and $22$ epochs respectively.

During testing, the images are normalized the same as the training process. As it is discovered that most of the foreground regions in the validation set are contained in the remaining top $300$ proposals after training, we only select the remaining top $300$ proposals as the input of Fast R-CNN in the testing stage. The $\sigma$ of both Soft-NMS and Embedding-Guided NMS is set to $0.5$. Same as the original setting of Faster R-CNN, only the top $100$ detections are reported as the final predictions. All experiments are conducted under a Ubuntu OS server with an Nvidia GTX Titan Xp GPU.

\subsection{Evaluation Results}

\begin{table*}
	\begin{small}
		\caption{Performance of DeepACEv2 in this paper on the testing set. The results are presented in all evaluation metrics. (FPN:Feature Pyramid Network, HNAS: Hard Negative Anchors Sampling, TM: Template Module, TNRL: Truncated Normalized Repulsion Loss, SOFT: Soft-NMS, EG: Embedding-Guided NMS)}\label{tabResults}
		\centering
		\begin{tabular*}{\textwidth}{@{}@{\extracolsep{\fill}}lcccccccccccccccccccc@{}}
			\toprule[2pt]
			Method &  WCR(\%) &  AER(\%) &  Acc(\%) &  $F_1$-score(\%)&  mAP(\%)&  $MR^{-2}$(\%)  \\
			\midrule[1pt]
			Faster R-CNN (ResNet-101+FPN) & 61.21 & 1.43 & 98.58  & 99.29 & 99.39  & \textbf{12.13} \\
			Faster R-CNN (ResNet-101+FPN+SOFT) & 68.48 & 1.22 & 98.79 & 99.39 & 99.58 & 18.99 \\
			\midrule
			DeepACEv2 (HNAS+TNRL+TM) & 66.02 & 1.31 & 98.69 & 99.34 & 99.44 & 13.29 \\
			DeepACEv2 (HNAS+TNRL+TM[SOFT]) & 70.67 & 1.18 & 98.82 & 99.41 & 99.58 & 15.24 \\
			DeepACEv2 (HNAS+TNRL+TM[EG])  & \textbf{71.39} & \textbf{1.17} & \textbf{98.84} & \textbf{99.42} & \textbf{99.60} & 14.52 \\
			\bottomrule[2pt]
		\end{tabular*}
	\end{small}
\end{table*}

\begin{figure*}
\centering
\includegraphics[width=\textwidth]{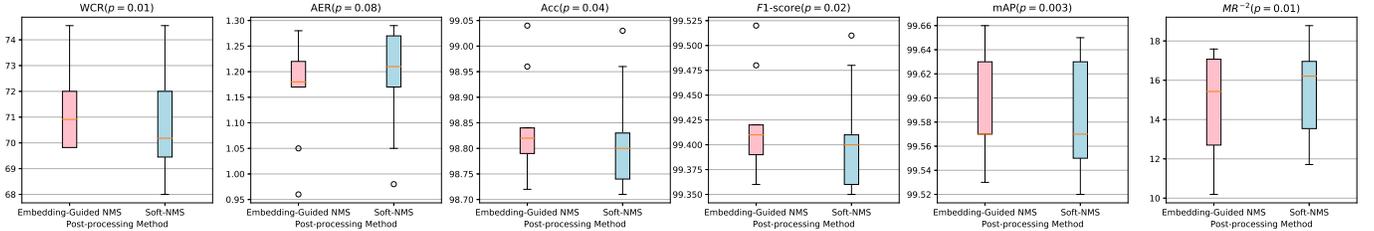}
\caption{We repeat experiments on testing set ten times and draw box plots for each metric. Meanwhile, the results of the statistical test (p-value) are shown in the head of the figure.}\label{p-value}
\end{figure*}

In this section, we provide the full evaluation results of the proposed methods on the testing set. Table \ref{tabResults} shows the performance of the Faster R-CNN, DeepACEv2, and their variants, which is evaluated by six metrics. Owing to the increasing development of advanced backbone and in-depth understanding of object detection problems, we take ResNet-101 as the backbone network of Faster R-CNN and attach the modified FPN after the backbone network to boost the performance. As shown in Table \ref{tabResults}, the new base framework achieves notable performances on WCR(\%) of $61.21$ and on AER, Acc, $F_1$-score, mAP, $MR^{-2}$(\%). Meanwhile, since Embedding-Guided NMS is inspired by Soft-NMS, we also evaluate the baseline advanced by Soft-NMS for a fair comparison. The performance is further boosted, and especially, WCR(\%) is increased by $7.27$.

\begin{figure}
\centering
\includegraphics[width=\columnwidth]{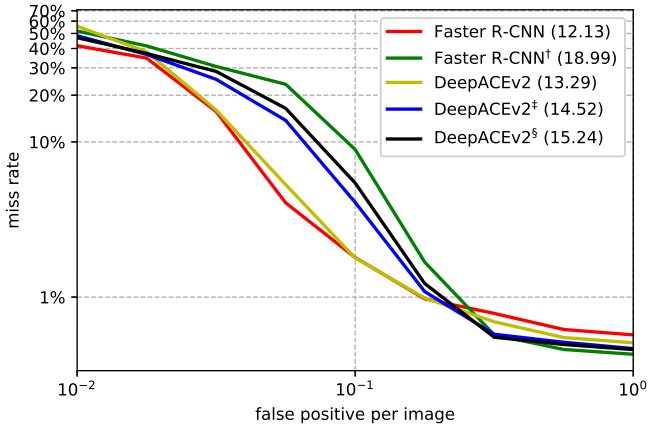}
\caption{Details about $MR^{-2}$ performance of Faster R-CNN, Faster R-CNN with Soft-NMS (Faster R-CNN$^\dag$), DeepACEv2, DeepACEv2 with Embedding-Guided NMS (DeepACEv2$^\ddag$) and DeepACEv2 with Soft-NMS (DeepACEv2$^\S$) on the testing set.}\label{mr}
\end{figure}

It is worth noting that the Soft-NMS method reduces the performance of $MR^{-2}$ of models. Specifically, Soft-NMS will increase the value of $MR^{-2}$(\%) for baseline from $12.13$ to $18.99$. The reason is because   that Soft-NMS has a higher miss rate than NMS at some specific FPPI (False Positive Per Image). Theoretically, Soft-NMS can have a lower overall miss rate since it does not delete the bounding box directly during post-processing but decay the score of it. However, a huge amount of true positives are slightly decayed simultaneously, which results in a higher miss rate when evaluating at lower FPPI, such as $10^{-2}$, as shown in Fig. \ref{mr}. As a result, the baseline equipped with Soft-NMS has poor performance on $MR^{-2}$.

Subsequently, Table \ref{tabResults} shows the performance of DeepACEv2 and its variants. DeepACEv2 that combines HNAS, Template Module, and TNRL improve WCR(\%) by $4.81$ comparing to the baseline of $61.21$. DeepACEv2 can also boost the performance of the model on the other four metrics except for $MR^{-2}$ lightly decreasing. However, as shown in Fig.\ref{mr}, the basic DeepACEv2 has a lower miss rate than basic Faster R-CNN except for strict FPPI condition, which means that DeepACEv2 can find more chromosomes although it slightly decrease the confidence scores on top detections. Meanwhile, with the implementation of Embedding-Guided NMS, DeepACEv2 also yields a performance boost comparing to DeepACEv2 with NMS and Soft-NMS. All six metrics are improved that are guaranteed by statistical tests, as shown in Fig.\ref{p-value}, and especially, WCR(\%) increases to $71.39$. Additionally, we can observe that DeepACEv2 with Embedding-Guided NMS has slightly poor performance on $MR^{-2}$ comparing to the basic Faster R-CNN. Same as the Soft-NMS, Embedding-Guided NMS also increases the miss rate on the condition of low FPPI. However, as illustrated in Fig.\ref{mr}, our proposed Embedding-Guided NMS can fix the shortage of Soft-NMS to retain more detection results during medium FPPI and final approach to the $MR^{-2}$ performance of basic Faster R-CNN. Especially, both three DeepACEv2 have better performances than baseline on the high FPPI conditions, indicating more chromosomes can be correctly detected. Overall, DeepACEv2 improves the performance by a large margin comparing to the baseline model.

\subsection{Comparison With Other Methods}
\begin{table*}
	\begin{small}
		\caption{The comparison of chromosomes counting methods on the testing set.}\label{tab1}
		\centering
		\begin{tabular*}{\textwidth}{@{}@{\extracolsep{\fill}}lccccccccccc@{}}
			\toprule[2pt]
			Method &  WCR(\%) & AER(\%) & $F_1$-score(\%) & mAP(\%)  \\
			\midrule[1pt]
			Gajendran et. al \cite{ref1} & 7.64 & 7.23 &- &-    \\ 
			Faster R-CNN (VGG16) &  39.64 & 2.44 & 98.79 & 99.03 \\
			DeepACEv1 &  47.63 & 2.39  & 98.81 & 99.45 \\
			DeepACEv2  &  \textbf{71.39} & \textbf{1.17} & \textbf{99.42} & \textbf{99.60}  \\
			\bottomrule[2pt]
		\end{tabular*}
	\end{small}
\end{table*}
In this section, we successively verify the effectiveness of our proposed method by comparing it with that of other methods. On the top of Table \ref{tab1}, we firstly show the chromosomes enumeration method proposed in \cite{ref1}, which is based on digital image analysis and evaluated on Metaphase Image Dataset and Background-Noise-Free Image Database. We reimplement and fine-tune this method on our collected dataset. However, since the old method counts chromosomes by searching for the skeleton of chromosomes rather than the bounding box regions, only the criterion of WCR and AER are comparable and summarized in Table \ref{tab1}.

Nevertheless, as shown in Table \ref{tab1}, our previously published method DeepACEv1 still dramatically outperforms the other method \cite{ref1}. Furthermore, it is worth mentioning that although previous work does not involve any pre-training in the detection head, it still significantly outperforms the Faster R-CNN that both detection head and VGG16 backbone have been pre-trained on the ImageNet datasets.

Owing to a powerful backbone network, advanced object detection toolbox, and enhanced methods, DeepACEv2 achieves the best performance comparing to previous works. Especially for WCR(\%), DeepACEv2 increases it by a large margin of $23.76$ compared to DeepACEv1. Moreover, DeepACEv2  achieves significant improvement on all the remaining metrics, in which it improves the relative values of AER by $51.05\%$, $F_1$-score by $51.26\%$, mAP by $27.27\%$, respectively.

\vspace{-5pt}
\subsection{Performance on touching and overlapping chromosomes}

As described in the introduction, chromosomes on metaphase images usually have severe touching and overlapping problems. In this section, we firstly describe the process of mapping a predicted bounding box to its corresponding ground truth, especially when overlapping and occlusion happens. Then, we define a criterion for severely touching or overlapping chromosomes and verify the performances of our method based on the subset of chromosomes.

\begin{figure}
\centering
\subfigure[]{\includegraphics[width=\columnwidth]{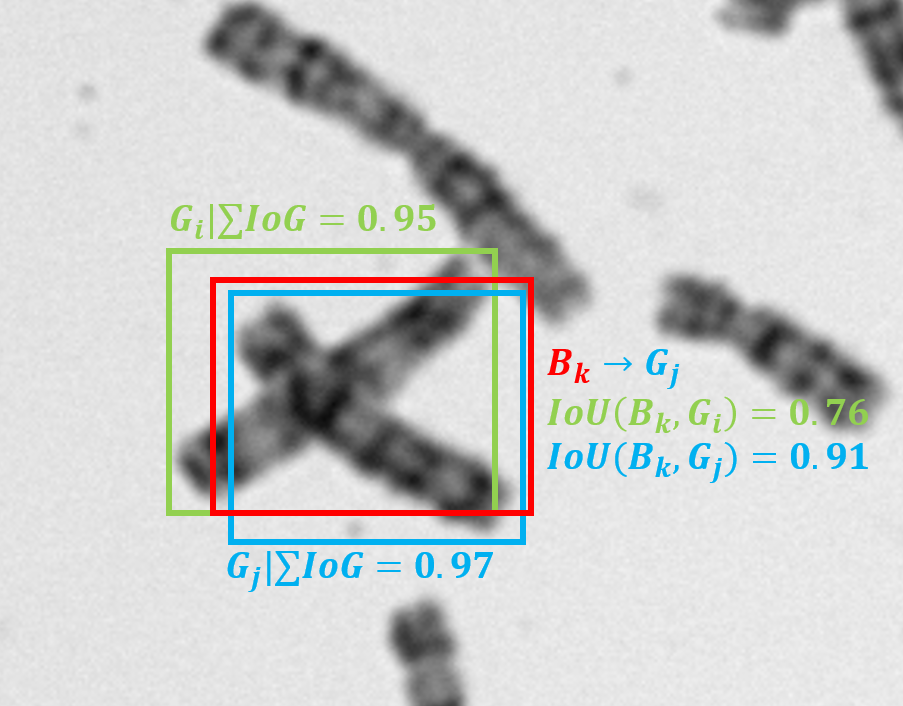}}
\quad
\subfigure[]{\includegraphics[width=0.45\columnwidth]{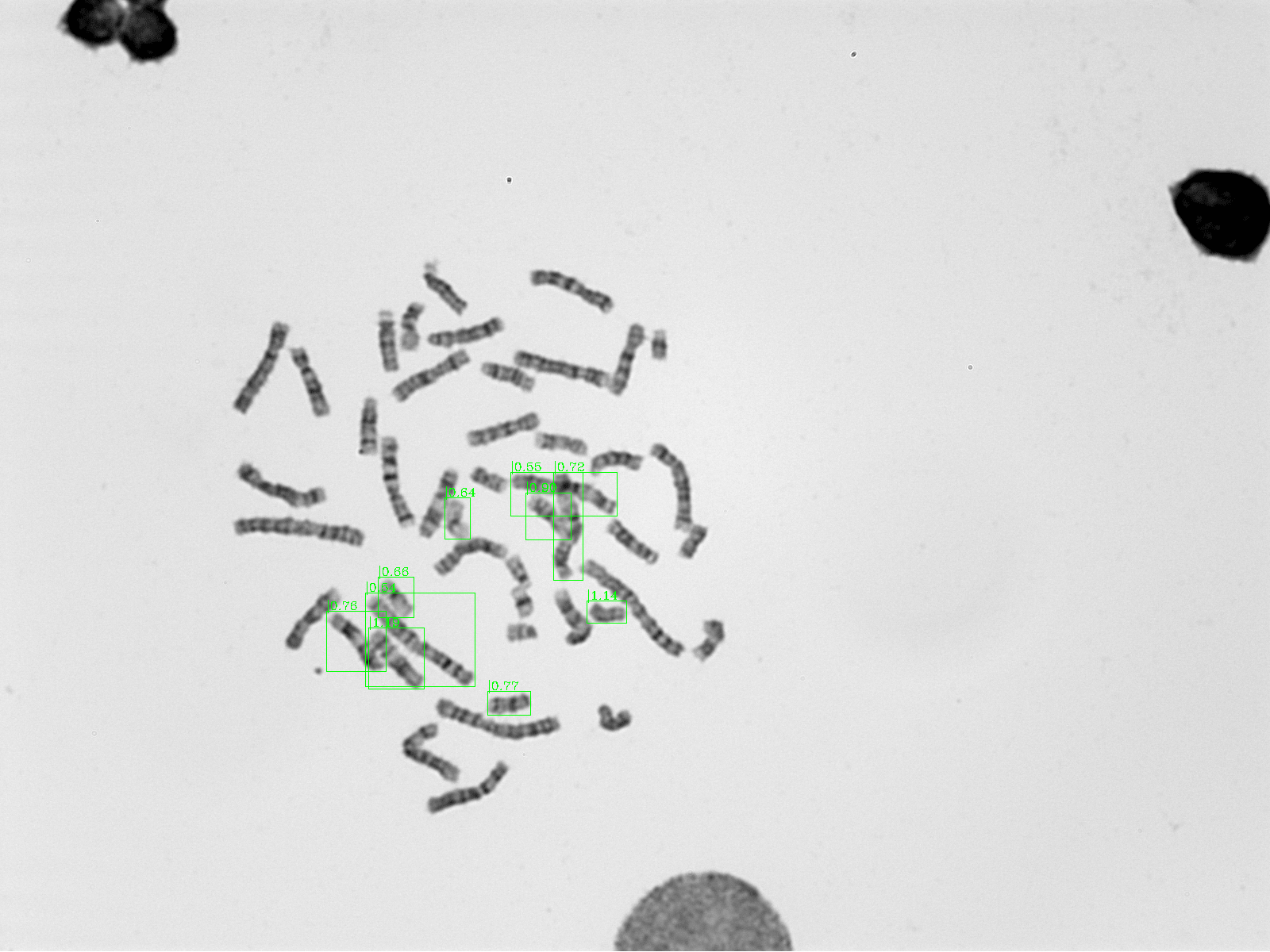}}
\quad
\subfigure[]{\includegraphics[width=0.45\columnwidth]{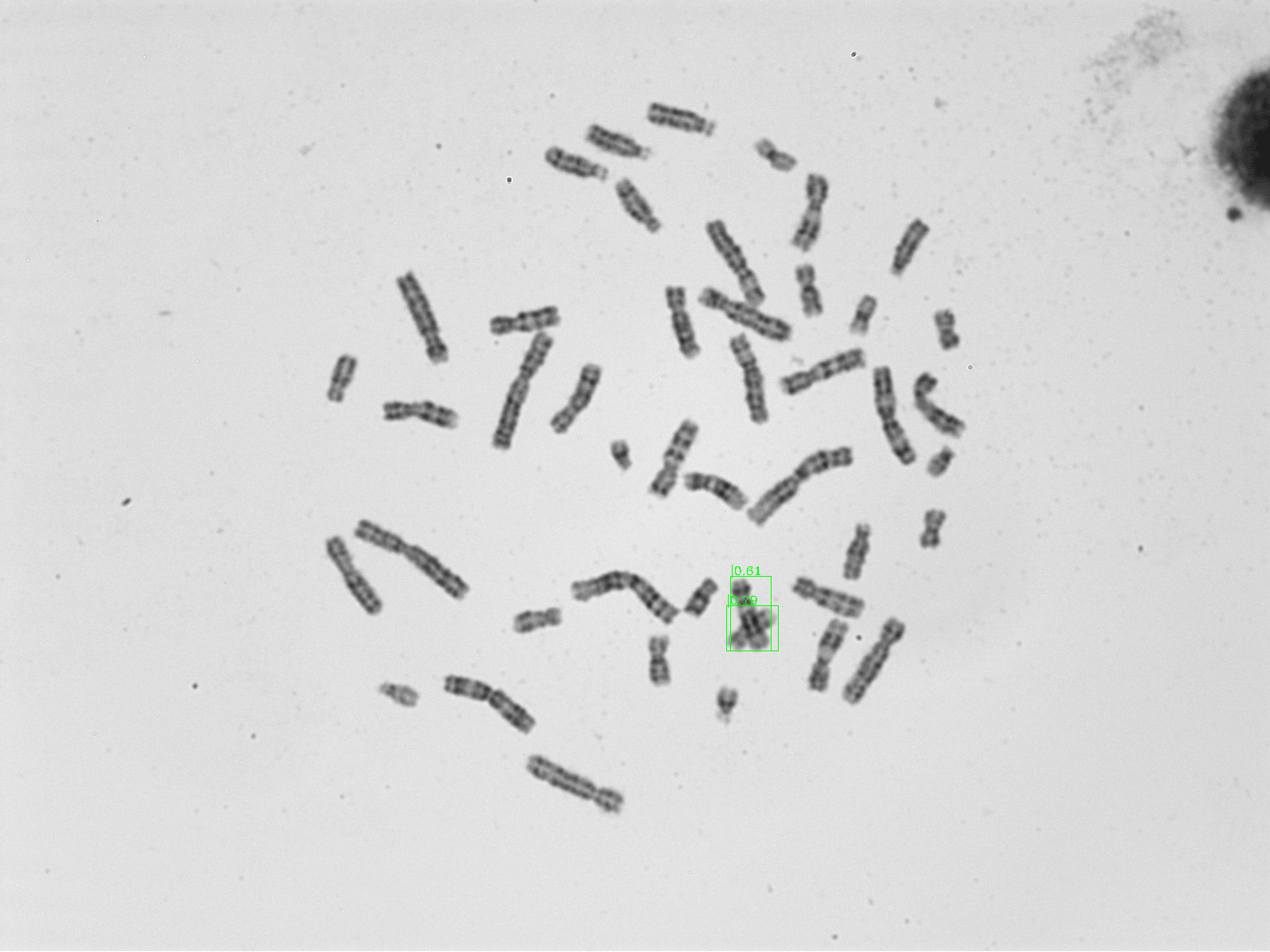}}
\caption{Illustration about the process of mapping a bounding box to its ground truth and select severe overlapping chromosomes. (a): green $G_i$ and blue $G_j$ rectangles are two ground truths, and the red $B_k$ rectangle is the predicted bounding box. Severely overlapping chromosomes set $\mathcal{S}_o$ includes $G_i$ and $G_j$ since both two ground truth have been occluded more than threshold 0.5. Besides, though $B_k$ are severe overlapped with both $G_i$ and $G_j$, we assign $B_k$ to $G_j$ rather than $G_i$ since $IoU(B_k, G_j) > IoU(B_k, G_i)$. (b) and (c) shows selected overlapping chromosomes on the metaphase images. All of the severely overlapped chromosomes are selected based on our criterion.}\label{overlap}
\end{figure}

\begin{table}[H]
	\caption{Statistics about overlapping chromosomes of each dataset.}\label{overlap_table}
	\centering
	\begin{tabular}{|l|c|c|c|}
	\hline
	 & training set & validation set & testing set  \\
	\hline
	 overlapping chromosomes & 3601 & 1110 & 1232 \\
	\hline
	 proportion  & 9.5\% & 8.8\% & 9.8\% \\
	 \hline
	\end{tabular}
\end{table}

\begin{table}[H]
	\caption{Performance of Faster R-CNN and DeepACEv2 on overlapping chromosomes of the testing set.}\label{overlap_result}
	\centering
	\begin{tabular}{|l|c|c|c|c|}
	\hline
						 & $F_1$-score(\%) & Precision(\%) & Recall(\%) & Acc(\%) \\
	\hline
	Faster R-CNN  &					96.84	&				97.93 &       95.78&    93.87\\
	\hline
	DeepACEv2		 &\textbf{97.93}&\textbf{98.77}&\textbf{97.10}&\textbf{95.94}\\
	\hline
	\end{tabular}
\end{table}

As illustrated in Fig. \ref{overlap}(a), to assign bounding box $B_k$, we need to compute $IoU(B_k, G_i)=\frac{B_k\cap G_i}{B_k \cup G_i}$ between $B_k$ and each ground truth $G_i$. Ground truths will be ranked according to IoU values and $B_k$ will be assignned to the ground truth which has the largest IoU with $B_k$:
\begin{equation}
B_k \to G_{\mathop{arg}\mathop{max}_i IoU(B_k, G_i)}
\end{equation}

Occlusion and overlapping of chromosomes are the most challenging problems for accurate detection of chromosomes. In our data set, all chromosomes are labeled by rectangle bounding boxes. Therefore, we can define the touching and overlapping chromosomes based on the interaction of bounding boxes. As shown in Fig. \ref{overlap}(a), the overlapping chromosomes subset $\mathcal{S}_o$ is defined as:

\begin{equation}
\mathcal{S}_o=\{G_i|\sum_{j=1,j\neq i}^nIoG(G_j,G_i) \ge \tau \}
\end{equation}
Here $IoG(G_j,G_i)\triangleq \frac{G_j \cap G_i}{G_i}$ has been described in Section \ref{TNRL} and $n$ is the number of ground truths in a metaphase image. $\tau$ is a predefined threshold, and we set $0.5$ in our work, which means that more than half an area of $G_i\in \mathcal{S}_o$ are overlapped with other bounding boxes as shown in Fig. \ref{overlap}. The statistics about overlapping chromosomes of each dataset is detailed in Table \ref{overlap_table}, nearly 10\% of chromosomes are severely overlapped with others.

Performances of the baseline and DeepACEv2 on the overlapping chromosomes are reported in Table \ref{overlap_result}. We evaluate Faster R-CNN and DeepACEv2 on overlapping chromosome subset $\mathcal{S}_o$ from the testing set. Comparing to the baseline, DeepACEv2 improves the $F_1$-score(\%) by $1.09$ and Acc(\%) by $2.07$. Especially, the great improvement of Recall value($+1.32$) proves that DeepACEv2 can alleviate the over deletion problem caused by severe overlapping chromosomes.

\subsection{Ablation Study}\label{Ablation}
\begin{table*}
	\setlength{\abovecaptionskip}{-10pt}
	\caption{Ablation Study of Different Component on Validation Set. All Experiments are repeated ten times, and mean values are reported.}\label{tabAblation}
	\centering
	\begin{tabular*}{\textwidth}{@{}@{\extracolsep{\fill}}lcccccc@{}}
		\toprule[2pt]
		Methods &                       WCR(\%) &AER(\%) & Acc(\%) & $F_1$-score(\%)& mAP(\%) & $MR^{-2}$(\%) \\  
		\midrule[1pt]
		Single1(ResNet-101+Single1-FPN) & 62.18 & 1.62   & 98.29   & 99.19   & 99.19   & 23.94 \\
		\midrule
		Single1+HNAS                    & 62.79 & 1.50   & 98.50   & 99.25   & 99.32   & 18.41 \\ 
		\midrule
		Single1+HNAS+TNRL               & 63.76 & 1.52   & 98.49   & 99.24   & 99.36   & \textbf{15.89} \\
		\midrule
		Single1+HNAS+TNRL+TM            & 63.42 & 1.51   & 98.50   & 99.25   & 99.34   & 16.83 \\
		Single1+HNAS+TNRL+TM[SOFT]      & 67.45 & 1.38   & 98.63   &99.31    & 99.40   & 18.32 \\
		Single1+HNAS+TNRL+TM[EG]        & \textbf{68.00} & \textbf{1.36} & \textbf{98.64} &\textbf{99.32}& \textbf{99.41}& 17.46 \\
		\midrule
		(EG vs SOFT) $p$-value          & 0.009 & 0.05 & 0.02 & 0.05 & 0.001 & 0.003 \\
		\bottomrule[2pt]
	\end{tabular*}
\end{table*}

\begin{figure}
	\centerline{\includegraphics[width=\columnwidth]{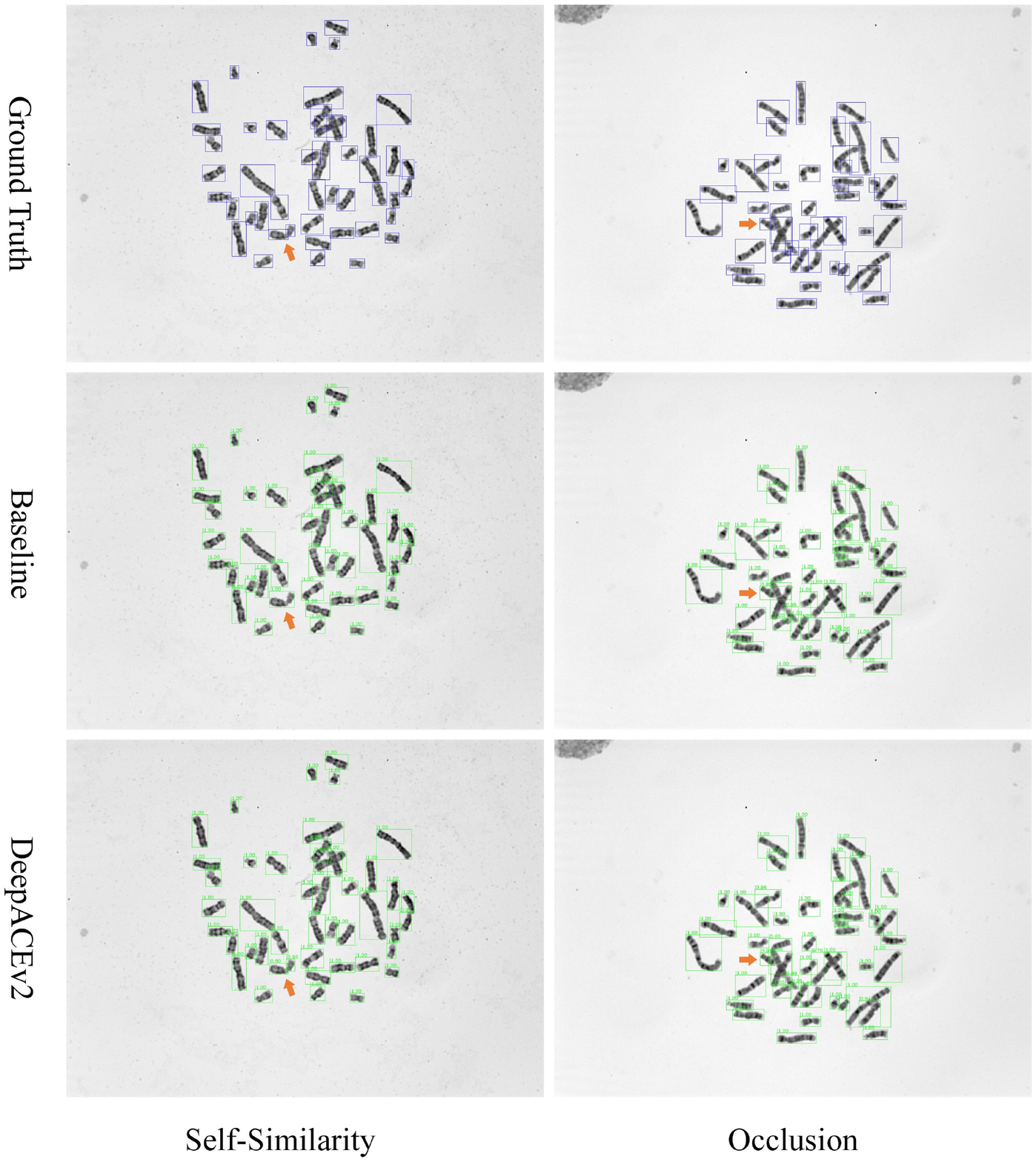}}
	\caption{The first row shows images with labeled ground truths, the second and third rows are prediction results of the baseline and DeepACEv2, respectively. Blue rectangular boxes represent labeled ground truth boxes, and green bounding boxes indicate predicted bounding boxes and the corresponding confidence scores. We use two typical examples to show the effectiveness of DeepACEv2.}
	\label{fig6}
\end{figure}

To justify the importance of each proposed module, Table \ref{tabAblation} summarizes the overall ablation studies. We add Hard Negative Anchors Sampling, Truncated Normalized Repulsion Loss, Template Module, and Embedding-Guided NMS on a single-level (as shown in Fig.\ref{fig1}.(a)) ResNet-101 FPN Faster R-CNN basic network step-by-step. For fair comparisons, experiments for ablation studies are kept identical with the final method except for specified changes in each ablation study. As shown in Fig. \ref{fig6}, DeepACEv2 is effective in solving the self-similarity and occlusion problems by adding the above three modules.

\subsubsection{Hard Negative Anchors Sampling} To verify HNAS's contribution to the performance, we firstly add HNAS to the basic network. Table \ref{tabAblation} shows that HNAS brings $0.61$ points higher WCR(\%) and $5.53$ points higher $MR^{-2}$ than the basic network. Simultaneously, HNAS can improve the relative values of the remaining four metrics by more than $7\%$. Noticed here that the basic network of DeepACEv2 is much more powerful than that used in DeepACEv1. However, the improvements here are also enough to validate the effectiveness of this module. Additionally, Table \ref{compare_HNAS} proves that the division criterion of HNAS used in this paper can achieve better performance than the settings in DeepACEv1.

\begin{table}[H]
	\caption{Comparison between two different settings of Hard Negative Anchors Sampling on the validation set: HNASv1 means the division criterion used in DeepACEv1 and HNASv2 means the division criterion used in this work.}\label{compare_HNAS}
	\centering
	\begin{tabular}{|l|c|c|c|c|c|c|}
	\hline
	Version & WCR & AER &  Acc& $F_1$-score& mAP& $MR^{-2}$ \\  
	\hline
	HNASv1 & 64.69 & 1.47 & 98.55 & 99.27 & \textbf{99.54} & 22.77 \\
	\hline
	HNASv2 & \textbf{68.00} & \textbf{1.36} & \textbf{98.64} & \textbf{99.32} & 99.41 & \textbf{17.46} \\
	\hline
	\end{tabular}
\end{table}

\subsubsection{Truncated Normalized Repulsion Loss}
Truncated Normalized Repulsion Loss improves the WCR(\%) from $62.73$ to $63.76$ and $MR^{-2}$(\%) from $19.13$ to $15.89$. To be more specific, as shown in Table \ref{mAP_TNRL}, combining the model with TNRL can achieve higher value in both $\text{mAP}_{50}$ and $\text{mAP}_{75}$. These results validate that TNRL can suppress the bounding box shifting and improve the localization accuracy of the model. However, TNRL may have slightly negative effects on the original detection loss, which may lead to some metrics slightly decreasing. In the following, we will use the Template Module to fix this tiny gap. 

\begin{table}[H]
	\caption{Mean mAP results in High IoU Threshold. Experiments are performed three times, and $\delta$ is the standard deviation.}\label{mAP_TNRL}
	\centering
	\begin{tabular}{|l|c|c|c|c|}
		\hline
		& $\text{mAP}_{50}$(\%) & $\delta$ & $\text{mAP}_{75}$(\%) & $\delta$ \\
		\hline
		w/o TNRL& 				99.32   &  0.04 &       98.57         & 0.07 \\
		w TNRL  &	   \textbf{99.35}   & 0.02   &  \textbf{98.67} & 0.04  \\
		\hline
	\end{tabular}
\end{table}

\subsubsection{Template Module with Embedding-Guided NMS}\label{TM}
As shown in Table \ref{tabAblation}, the Template Module with Embedding-Guided NMS improves the performance significantly. The combination of them improves the WCR(\%) from $63.76$ to $68.00$ and remaining AER, Acc, $F_1$-score, mAP, and $MR^{-2}$ also have been greatly improved. Besides, we also compare the Template Module designed and used in DeepACEv2 with the one used in DeepACEv1 \cite{ref43}, and results in Table \ref{tabTMv1v2} show that the new Template Module can achieve better performance. It is interesting to notice that (as shown in Table \ref{tabTM}), the Embedding-Guided NMS achieves better performance by slightly sacrificing precision while increasing recall compared to Soft-NMS. Furthermore, the Embedding-Guided NMS also recovers the performance reduction of $MR^{-2}$ brought by the Soft-NMS module. Finally, statistical test results in Table \ref{tabAblation} prove that Embedding-Guided NMS can improve the performance comparing to Soft-NMS, which validates the importance of embeddings in the post-processing procedure. 

\begin{table}[H]
\caption{Comparison between the Template Modules used in DeepACEv1 and DeepACEv2 on validation set: TMv1 means the Template Module used in DeepACEv1 and TMv2 means the Template Module used in DeepACEv2.  All experiments are repeated ten times, and mean values are reported.}\label{tabTMv1v2}
\centering
	\begin{tabular}{|l|c|c|c|c|c|c|}
	\hline
	Version & WCR & AER &  Acc& $F_1$-score& mAP& $MR^{-2}$ \\  
	\hline
   TMv1 & \textbf{64.00} & 1.56 & 98.45 & 99.22 & 99.25 & 20.08 \\
	\hline
	TMv2 & 63.42 & \textbf{1.51} & \textbf{98.50} & \textbf{99.25}   & \textbf{99.34} & \textbf{16.83} \\
	\hline
	\end{tabular}
\end{table}

\begin{table}[H]
	\caption{The Precision and Recall of different post-processing methods, SOFT(Soft-NMS) vs. EG(Embedding-Guided NMS).  All experiments are repeated ten times, and mean values are reported. }\label{tabTM}
	\centering
	\begin{tabular}{|l|c|c|c|c|}
		\hline
		& Precision(\%) & $p$-value &  Recall(\%) & $p$-value \\
		\hline
		w SOFT             &\textbf{99.50} & \multirow{2}{*}{0.02} &   99.12 & \multirow{2}{*}{0.0004}   \\
		w EG &        99.49  & \multicolumn{1}{c|}{}       &\textbf{99.15} & \multicolumn{1}{c|}{}    \\
		\hline
	\end{tabular}
\end{table}

\section{Conclusion}\label{Conclusion}
In this paper, we develop an automated chromosome enumeration algorithm with higher performance, DeepACEv2. A Hard Negative Anchors Sampling strategy is adopted to learn more about partial chromosomes. Template Module equipped with Embedding-Guided NMS inspired by associative embedding mechanism is designed to identify overlapping chromosomes heuristically. To alleviate serious occlusion problems, we novelly design the Truncated Normalized Repulsion Loss to avoid bounding box regression error when occlusion happens. Experiments on clinical datasets demonstrate its effectiveness. The future plan is to continue to develop methods to solve chromosomes classification and segmentation tasks based on the whole metaphase images. 

\section{Acknowledgement}
We are grateful to the anonymous reviewers for their helpful comments. We thank Professor S. Kevin Zhou for providing critical comments, and Yuwei Xiao at the Carnegie Mellon University to help proofreading during manuscript preparation.

\section{Author Contributions}
Tianqi Yu, Manqing Wang, Fuhai Yu, Chan Tian, and Jie Qiao collected and labeled the data. Li Xiao, Chunlong Luo, Yufan Luo and Yinhao Li designed the model and analyzed the data. Chunlong Luo implemented the model. Li Xiao conceived and supervised this work and wrote the manuscript with assistance from Jie Qiao and Chan Tian. Further information or questions should be directed to the Lead Contact, Li Xiao (xiaoli@ict.ac.cn). 

\bibliographystyle{IEEEtran}
\normalem
\bibliography{citation}
\end{document}